\documentclass{article}

\PassOptionsToPackage{numbers, compress}{natbib}

 \usepackage[main, final]{neurips_2026}

\usepackage[utf8]{inputenc} 
\usepackage[T1]{fontenc}    
\usepackage{hyperref}       
\usepackage{url}            
\usepackage{booktabs}       
\usepackage{amsfonts}       
\usepackage{nicefrac}       
\usepackage{microtype}      
\usepackage{xcolor}         

\usepackage{amsmath}
\usepackage{enumitem}
\usepackage{colortbl}
\usepackage{pifont} 
\usepackage{amssymb}
\usepackage{tcolorbox}
\tcbuselibrary{skins, breakable}

\definecolor{highlightrow}{RGB}{235, 248, 255} 
\definecolor{cgreen}{RGB}{50, 160, 80} 
\definecolor{cred}{RGB}{200, 50, 50}   
\definecolor{cgray}{RGB}{150, 150, 150} 

\newcommand{\yes}{\textcolor{cgreen}{\ding{51}}} 
\newcommand{\no}{\textcolor{cgray}{\ding{55}}}   

\usepackage{booktabs}
\usepackage{multirow}
\usepackage{colortbl}
\usepackage{xcolor}
\usepackage{pifont}
\usepackage{tabularx} 
\usepackage{makecell}
\definecolor{rowblue}{RGB}{242, 247, 255} 
\definecolor{myteal}{RGB}{45, 160, 160}   
\definecolor{mygray}{RGB}{190, 190, 190}  

\def\eg{\emph{e.g.}} 
\def\ie{\emph{i.e.}} 

\definecolor{headergray}{RGB}{245, 245, 245}   
\definecolor{bestblue}{RGB}{0, 50, 150}        
\definecolor{textgray}{RGB}{120, 120, 120}     

\newtcolorbox{promptbox}[1]{
    enhanced,
    colback=gray!5!white,      
    colframe=gray!60!black,    
    coltitle=white,            
    title=\textbf{\small \textsc{#1}}, 
    fontupper=\footnotesize\ttfamily, 
    boxrule=0.8pt,
    arc=2pt,
    left=6pt, right=6pt, top=6pt, bottom=6pt,
    breakable,                
    skin=enhanced,
}

\newcommand{\appitem}[2]{%
  \noindent\textbf{\hyperref[#1]{#2}}\hfill\textbf{\pageref{#1}}\par\vspace{0.55em}
}

\newcommand{\appsubitem}[2]{%
  \noindent\hspace{1.8em}\hyperref[#1]{#2}\dotfill\pageref{#1}\par
}
\title{Can Vision-Language Models Think from the Sky? Unifying UAV Reasoning and Generation}

\author{
  Jintao Sun \\
  \small Beijing Institute of Technology \\
  {\small\texttt{3120215524@bit.edu.cn}}
  \And
  Gangyi Ding \\
  \small Beijing Institute of Technology \\
  {\small\texttt{dgy@bit.edu.cn}}
  \And
  Donglin Di \\
  \small Harbin Institute of Technology \\
  {\small\texttt{donglin.ddl@gmail.com}}
  \AND
  Hu Zhang \\
  \small CSIRO Data61 \\
  {\small\texttt{hu1.zhang@csiro.au}}
  \And
  Zhedong Zheng\thanks{Corresponding author.} \\
  \small University of Macau \\
  {\small\texttt{zhedongzheng@um.edu.mo}}
}

\begin{document}
\maketitle

\vspace{-0.7cm}
\begin{abstract} 
Vision-Language Models have achieved strong progress in ground-view visual understanding, yet they remain brittle in high-altitude Unmanned Aerial Vehicle scenes, where objects are tiny and densely packed, textures are repetitive, and top-down orientations are ambiguous. We introduce UAVReason, a large-scale UAV-native dataset and evaluation suite for studying unified aerial reasoning and generation under this nadir-view domain shift. UAVReason aligns RGB imagery, depth maps, semantic segmentation masks, captions, and question-answer pairs within a consistent aerial domain. It contains 23.6K captioned frames, 273K VQA pairs including 68.2K two-frame temporal questions, and 188.8K cross-modal generation samples across RGB, depth, and segmentation modalities. We further adapt UAVReason-Bagel as a unified understanding-and-generation baseline that jointly optimizes language reasoning and dense visual generation objectives. Experiments show that general-purpose VLMs and off-the-shelf unified generators struggle with UAV-native grounding, while UAVReason-Bagel substantially improves over its pretrained counterpart, increasing VQA-1F F1 from 0.394 to 0.711, VQA-2F F1 from 0.427 to 0.822, and heading-aware VQA F1 from 0.798 to 0.973. For generation, it improves segmentation mIoU to 0.143 and reduces KID from 0.078 to 0.048 for depth-segmentation-text-conditioned RGB synthesis. More importantly, our ablations reveal a bidirectional synergy between synthesis and reasoning. Dense generation objectives improve temporal semantic consistency, while language-level reasoning regularizes sparse-condition image synthesis. These results suggest that unified reasoning and generation provide effective geometry-aware structural priors for physically grounded aerial intelligence. All data, code, and evaluation tools will be released.
\end{abstract}

\begin{figure}[ht]
    \centering
    \vspace{-1.2cm}
    \includegraphics[width=0.98\linewidth]{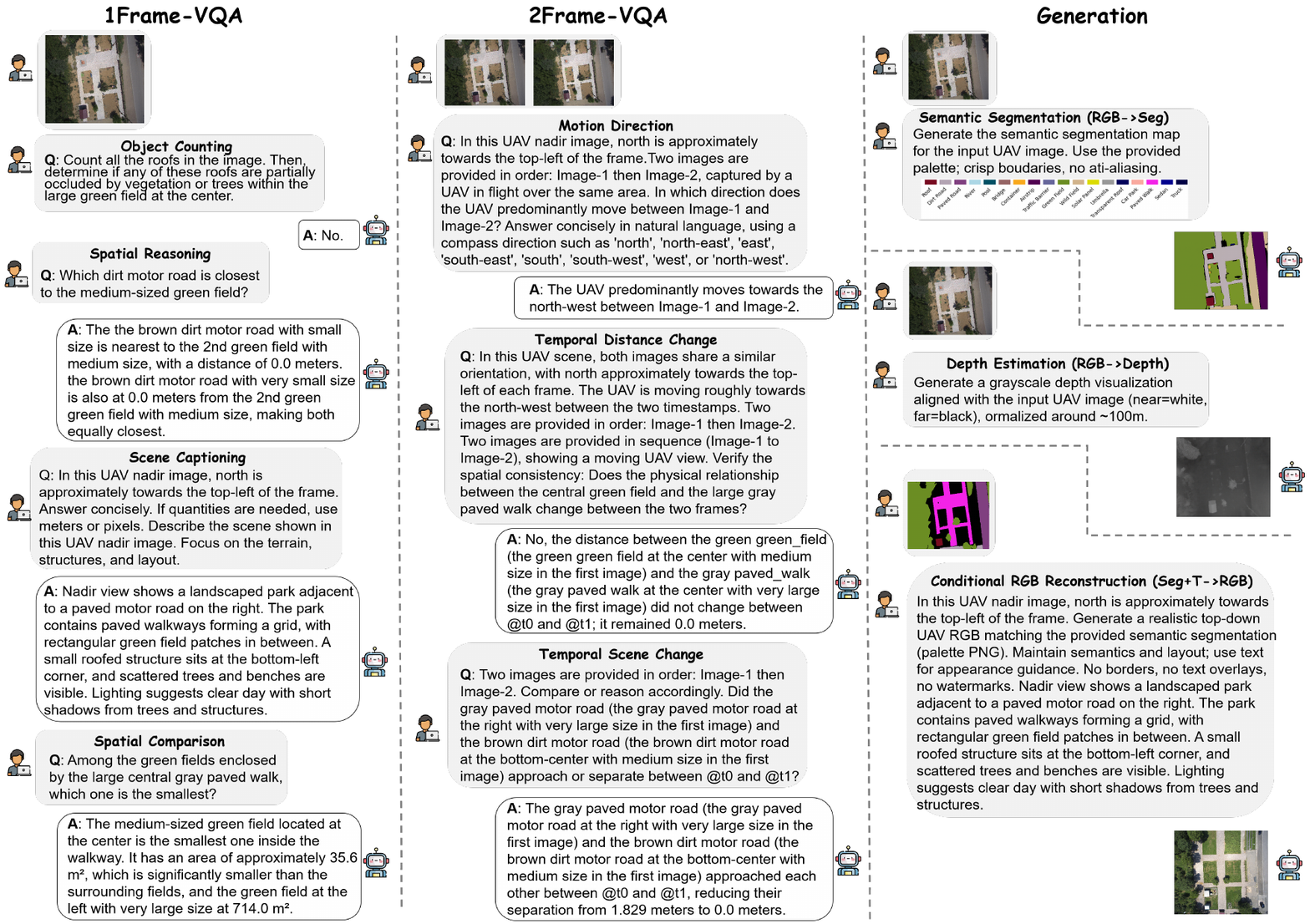}
    \vspace{-0.5em}
    \caption{\textbf{UAVReason}, a unified benchmark for nadir-view spatio-temporal reasoning and cross-modal generation.
    \textbf{1Frame-VQA (Left):} Single-frame reasoning with orientation cues (\eg, north), requiring object counting, scene captioning, spatial reasoning, and comparison ability.
    \textbf{2Frame-VQA (Middle):} Two-frame temporal reasoning over aligned viewpoints, probing motion direction and temporal changes in relative distance and scene changes.
    \textbf{Generation (Right):} Pixel-level conditional synthesis with diverse inputs, covering semantic segmentation (RGB$\rightarrow$Seg), depth estimation (RGB$\rightarrow$Depth), and conditional RGB reconstruction from segmentation and text prompts~(Seg+T$\rightarrow$RGB).
    }
    \label{fig:1}
    \vspace{-1.5em}
\end{figure}

\section{Introduction} \label{sec:introduction}
Vision-Language Models (VLMs) have reshaped visual understanding by aligning images and text at scale, enabling open-ended recognition, grounding, and reasoning across diverse tasks~\citep{clip,gpt4v,llava,SpatialRGPT}. Such capabilities are increasingly central to embodied agents, including Unmanned Aerial Vehicles (UAVs), mobile robots, and autonomous systems, that perceive, reason, and act in complex, dynamic environments. 
However, today’s VLMs remain largely ground-centric and often degrade in aerial scenarios, since most pretraining corpora are dominated by human-centric, ground-view imagery~\citep{coco,f30k}. When deployed on high-altitude UAV platforms, models face a pronounced domain shift: objects appear tiny and densely packed; textures are repetitive (\eg, rooftops and foliage); and the absence of a stable horizon yields ambiguous top-down orientation. 


As summarized in Table~\ref{tab:dataset_comparison_final}, existing aerial datasets primarily focus on individual tasks such as detection~\citep{visdrone_2021,dota}, tracking and 3D perception~\citep{du2018unmanned,UAV3D}, and segmentation~\citep{uavid,aeroscapes,isaid}. While recent efforts explore monocular depth~\citep{pan2024mona,wilduav}, embodied intelligence in urban spaces~\citep{urbanvideobench_2025}, or dedicated UAV VQA~\citep{hrvqa,earthvqa}, they typically lack integrated geometric supervision and cross-modal generation within a unified protocol. 
This fragmentation results in isolated training of understanding and generation tasks. 
Consequently, models fail to benefit from the complementary geometric cues shared across tasks, which are crucial for learning robust representations in nadir-view UAV scenarios.
In this paper, we argue that UAV-native capabilities require jointly modeling ``reasoning'' and ``generation''.
Perceiving small, ambiguous overhead objects requires precise pixel-level spatial reasoning beyond semantic alignment. 
For models that can answer VQA questions, those that can additionally generate segmentation maps from text exhibit a richer form of scene understanding: they capture not only what an object is, but also where it is and what its spatial shape looks like.




To fill this gap, we introduce \textbf{UAVReason}, a unified large-scale nadir-view benchmark for \emph{multimodal UAV scene reasoning and generation} (see Figure~\ref{fig:1}). Built from a single, high-fidelity UAV platform to ensure domain consistency, UAVReason consolidates dense RGB–Depth–Segmentation supervision with two complementary tracks: \textbf{Reasoning Track}: over \textbf{273,000} VQA pairs including \textbf{23,600} single frames each with detailed scene captions, and \textbf{68,223} temporal 2-Frame sequences. \textbf{Generation Track}: \textbf{188,800} cross-modal samples spanning tasks, \eg, RGB$\rightarrow$Segmentation, Depth/Segmentation$\rightarrow$RGB, and text-conditioned generation.  Our main contributions are as follows:

\begin{itemize}
    \item \textbf{UAV-Native Evaluation Suite.}
    We introduce UAVReason, a large-scale unified evaluation suite that aligns RGB imagery, dense geometry including depth and segmentation, and language annotations to study multimodal models under nadir-view domain shift.

    \item \textbf{Unified Aerial Protocol.}
    UAVReason integrates 23.6K captioned frames, 273K VQA pairs including 68.2K two-frame temporal questions, and 188.8K cross-modal generation samples, enabling joint evaluation of 22 spatio-temporal reasoning types and geometrically grounded visual synthesis.

    \item \textbf{Reasoning and Generation Synergy.} We establish UAVReason-Bagel as a unified understanding-and-generation baseline and use it to examine the interaction between synthesis and reasoning, showing that dense generation objectives improve temporal semantic consistency while language-level reasoning regularizes sparse-condition image synthesis.
\end{itemize}
\begin{table*}[t]
\centering
\vspace{-.5in}
\scriptsize
\setlength{\tabcolsep}{4.2pt} 
\renewcommand{\arraystretch}{1.2}
\caption{\textbf{Comparison with existing UAV, aerial, and remote sensing benchmarks.}
We summarize representative datasets by category and by supported supervision and tasks, including meta properties (Typ, Vw), dense geometry signals (Dep, Seg), reasoning capabilities (VQA, Tmp, Cap), and generation support (CMG), together with dataset scale and statistics.
\textit{Legend:} \textbf{Typ} denotes real (R) or synthetic (S); \textbf{Vw} denotes nadir (N), oblique (O), or mixed (M) views; \textbf{Dep} and \textbf{Seg} indicate dense depth and segmentation annotations; \textbf{VQA}, \textbf{Tmp}, and \textbf{Cap} indicate visual question answering, temporal reasoning, and captioning; \textbf{CMG} indicates cross-modal generation.}
\begin{tabularx}{\textwidth}{l l c c c c c c c c c X} 
\toprule

\multirow{3.5}{*}{\textbf{Category}} & 
\multirow{3.5}{*}{\textbf{Dataset}} & 
\multicolumn{2}{c}{\textbf{Meta}} & 
\multicolumn{2}{c}{\textbf{Dense Geom.}} & 
\multicolumn{3}{c}{\textbf{Reasoning}} & 
\multicolumn{1}{c}{\textbf{Gen.}} & 
\multirow{3.5}{*}{\textbf{Scale \& Stats}} \\

\cmidrule(lr){3-4} \cmidrule(lr){5-6} \cmidrule(lr){7-9} \cmidrule(lr){10-10}

& & \textbf{Typ} & \textbf{Vw} & 
\textbf{Dep} & \textbf{Seg} & 
\textbf{VQA} & \textbf{Tmp} & \textbf{Cap} & 
\textbf{CMG} & \\

\midrule

\multirow{5}{*}{\shortstack[l]{\textit{Detection /}\\\textit{Segmentation}}} 
& VisDrone~\citep{visdrone_2021} & R & M & \no & \no & \no & \no & \no & \no & 262k frames, 10k imgs \\
& UAVDT~\cite{du2018unmanned} & R & M & \no & \no & \no & \no & \no & \no & $\sim$80k frames \\
& UAVid~\citep{uavid} & R & O & \no & \yes & \no & \no & \no & \no & 300 labeled 4K seqs \\
& AeroScapes~\citep{aeroscapes} & R & M & \no & \yes & \no & \no & \no & \no & 3.3k images \\
& DOTA/iSAID~\citep{dota} & R & M & \no & \yes & \no & \no & \no & \no & 188k/655k instances \\
\midrule

\multirow{3}{*}{\shortstack[l]{\textit{Geometry /}\\\textit{Multi-sensor}}} 
& UAVScenes~\citep{uavscenes2025} & R & M & \yes & \yes & \no & \no & \no & \no & $>$120k frames (RGB+L) \\
& WildUAV~\citep{wilduav} & R & M & \yes & \no & \no & \no & \no & \no & $>$1.5k high-res imgs \\
& MoNA Bench~\cite{pan2024mona} & R & M & \yes & \no & \no & \no & \no & \no & MAV nav. benchmarks \\
\midrule

\multirow{4}{*}{\textit{Synthetic}} 
& Mid-Air~\cite{9025697} & S & M & \yes & \yes & \no & \no & \no & \no & Sim. trajectories \\
& TartanAir~\cite{9341801} & S & M & \yes & \yes & \no & \no & \no & \no & Sim. navigation \\
& SynDrone~\cite{10350525} & S & M & \yes & \yes & \no & \no & \no & \no & Physics simulation \\
& University-1652~\cite{zheng2020university} & R & M & \no & \no & \no & \no & \no & \no & 72 univs / 1652 bldgs; Google Earth \\

\midrule

\multirow{7}{*}{\textit{VQA / RS}} 
& RSVQA-LR~\citep{RSVQA}  & R & N & \no & \no  & \yes & \no & \no  & \no & 772 imgs, 77k QA \\
& RSVQA-HR~\citep{RSVQA}  & R & N & \no & \no  & \yes & \no & \no  & \no & 10.7k imgs, 1.07M QA \\
& RSICD~\cite{RSICD}     & R & M & \no & \no  & \no  & \no & \yes & \no & 10.9k imgs, 5 Cap/img \\
& HRVQA\citep{hrvqa} & R & N & \no & \no & \yes & \no & \no & \no & 53k imgs, 1M QA \\
& EarthVQA~\citep{earthvqa} & R & N & \no & \yes & \yes & \no & \no & \no & 6k imgs, 208k QA \\
& RSIVQA~\citep{rsivqa} & R & M & \no & \no & \yes & \no & \no & \no & 37k imgs, 111k QA \\
& VRSBench~\cite{vrsbench} & R & M & \no & \no & \yes & \no & \yes & \no & 29k imgs, Cap+QA \\
\midrule

\rowcolor{highlightrow} 
\textbf{Unified} & \textbf{UAVReason} & \textbf{R} & \textbf{N} & 
\textbf{\yes} & \textbf{\yes} & 
\textbf{\yes} & \textbf{\yes} & \textbf{\yes} & 
\textbf{\yes} & 

\textbf{\makecell[c]{23.6k frames with Caps; 68.2k Seq\\ 273k VQA; 188.8k Gen Pairs}} \\

\bottomrule
\end{tabularx}
\label{tab:dataset_comparison_final}
\vspace{-.15in}
\end{table*}
\section{Related Work} \label{sec:related_work}
\vspace{-.05in}
\noindent\textbf{UAV Vision Benchmarks.} 
The UAV aerial imagery landscape has evolved from detection-oriented benchmarks, \eg, VisDrone~\citep{visdrone_2021} and DOTA~\citep{dota}, to tracking and 3D perception benchmarks~\citep{du2018unmanned,UAV3D,bucktales}, dense prediction datasets, \eg, UAVid~\citep{uavid}, and high-level reasoning, \eg, HRVQA~\citep{hrvqa}. However, these datasets remain largely task-specific, isolating perception from reasoning and lacking integrated evaluation of spatio-temporal VQA with pixel-level generation. In contrast, UAVReason overcomes these limitations by unifying perception, reasoning, and generative capabilities on a consistent nadir-view platform. 

\noindent\textbf{Vision-Language Models.} 
Standard VLMs, \eg, LLaVA-OneVision~\citep{llava-onevision} and Qwen2-VL~\citep{qwen2.5-vl}, achieve remarkable general-purpose understanding~\citep{naturalbench,mmstar}. However, they suffer from pronounced ground-centric bias when transferred to nadir views.
More critically, their discriminative nature restricts outputs to textual responses, thereby hindering the pixel-level synthesis (\eg, depth/segmentation) required for dense spatial grounding in aerial environments.
To transcend text-only outputs, recent works, \ie, Janus~\citep{janus}, OmniGen~\citep{omnigen}, and Bagel~\citep{BAGEL} have integrated generative capabilities into transformer-based models. 
Despite their promise, these general-domain models lack supervision from paired dense geometry under aerial viewpoints. Consequently, they remain vulnerable to hallucinations and unstable grounding when deployed in aerial scenarios.

\noindent\textbf{Aerial-adapted VLMs.} 
Recently, some approaches, \eg, GeoChat~\citep{geochat} and EarthGPT~\citep{earthgpt} mitigate the domain shift via instruction tuning but lack support for controllable cross-modal generation, limiting their ability to enforce structural consistency.
In contrast, we fine-tune a unified architecture on UAVReason, establishing a baseline that couples reasoning with pixel-level synthesis. We show that incorporating generative objectives provides effective structural priors, leading to improved spatial perception.
\vspace{-.15in}
\section{UAVReason Benchmark}
\vspace{-.05in}
\label{sec:UAVReason}

\subsection{Dataset Construction}
\label{sec:dataset_construction}

\begin{figure*}[t!]
  \centering
  \vspace{-.3in}
  \includegraphics[width=\linewidth]{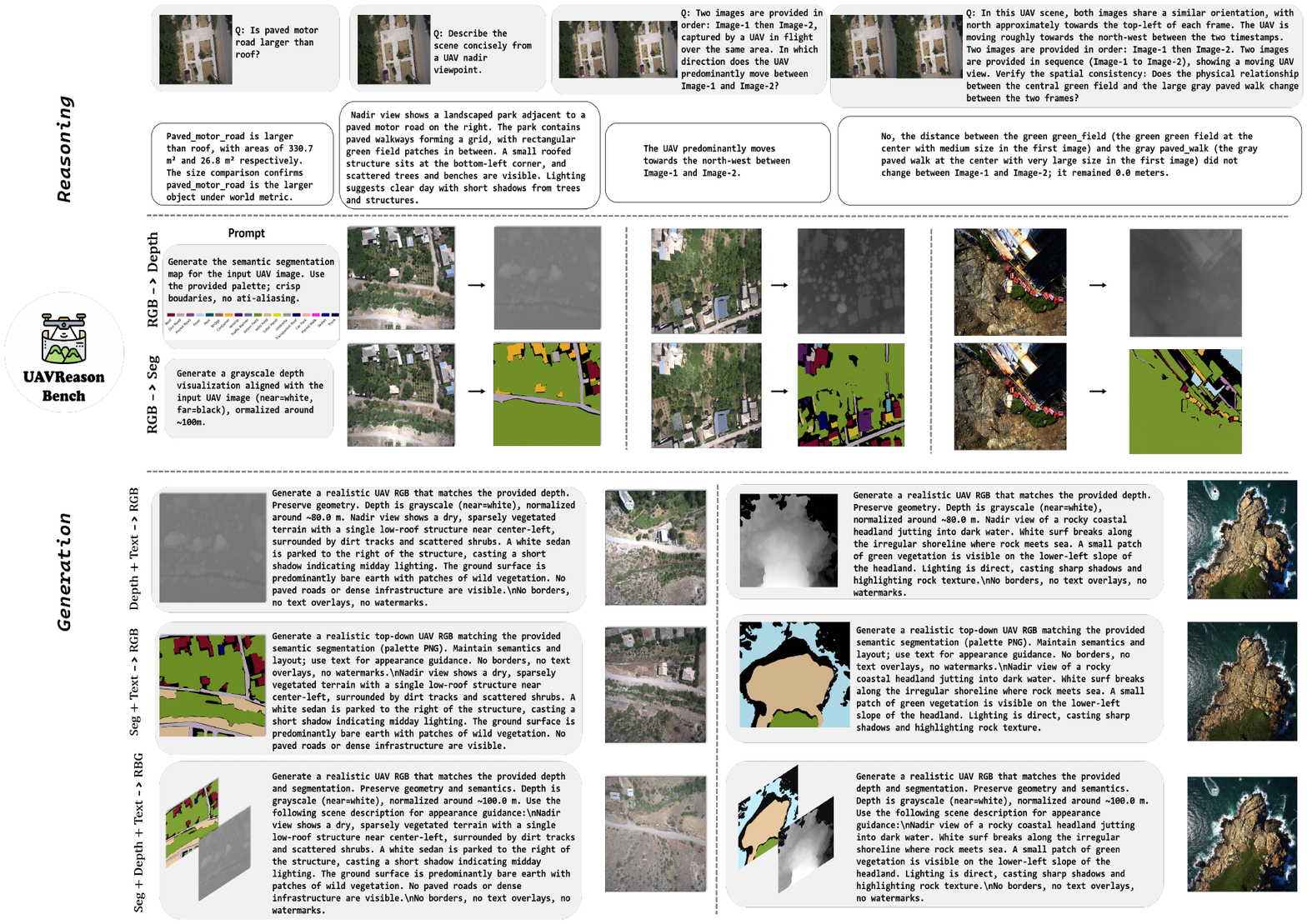}
  \vspace{-.2in}
    \caption{\textbf{Overview of UAVReason benchmark tasks and I/O protocols.}
    \textbf{Top (Reasoning):} Language-centric reasoning on nadir-view UAV imagery, including single-frame spatial queries (\eg, counting/comparison and referring questions with orientation cues), global scene captioning, and two-frame temporal reasoning (motion direction and cross-time relational verification).
    \textbf{Middle (Geometry prediction):} Dense pixel-level supervision aligned to the same UAV frames, covering RGB$\rightarrow$Depth (grayscale depth visualization, near-to-far) and RGB$\rightarrow$Seg (palette PNG with crisp boundaries).
    \textbf{Bottom (Conditional generation):} Controllable RGB synthesis conditioned on geometric modalities and text, including Depth+Text$\rightarrow$RGB, Seg+Text$\rightarrow$RGB, and Depth+Seg+Text$\rightarrow$RGB.
    Together, these tasks unify semantic reasoning with geometric/structural constraints, supporting unified training and evaluation of grounding fidelity, geometric consistency, and controllable synthesis in the UAV domain.}
    \label{fig:all_task}
\vspace{-.2in}
\end{figure*}

\begin{figure}[t!]
  \centering
\vspace{-.2in}
  \includegraphics[width=0.7\linewidth]{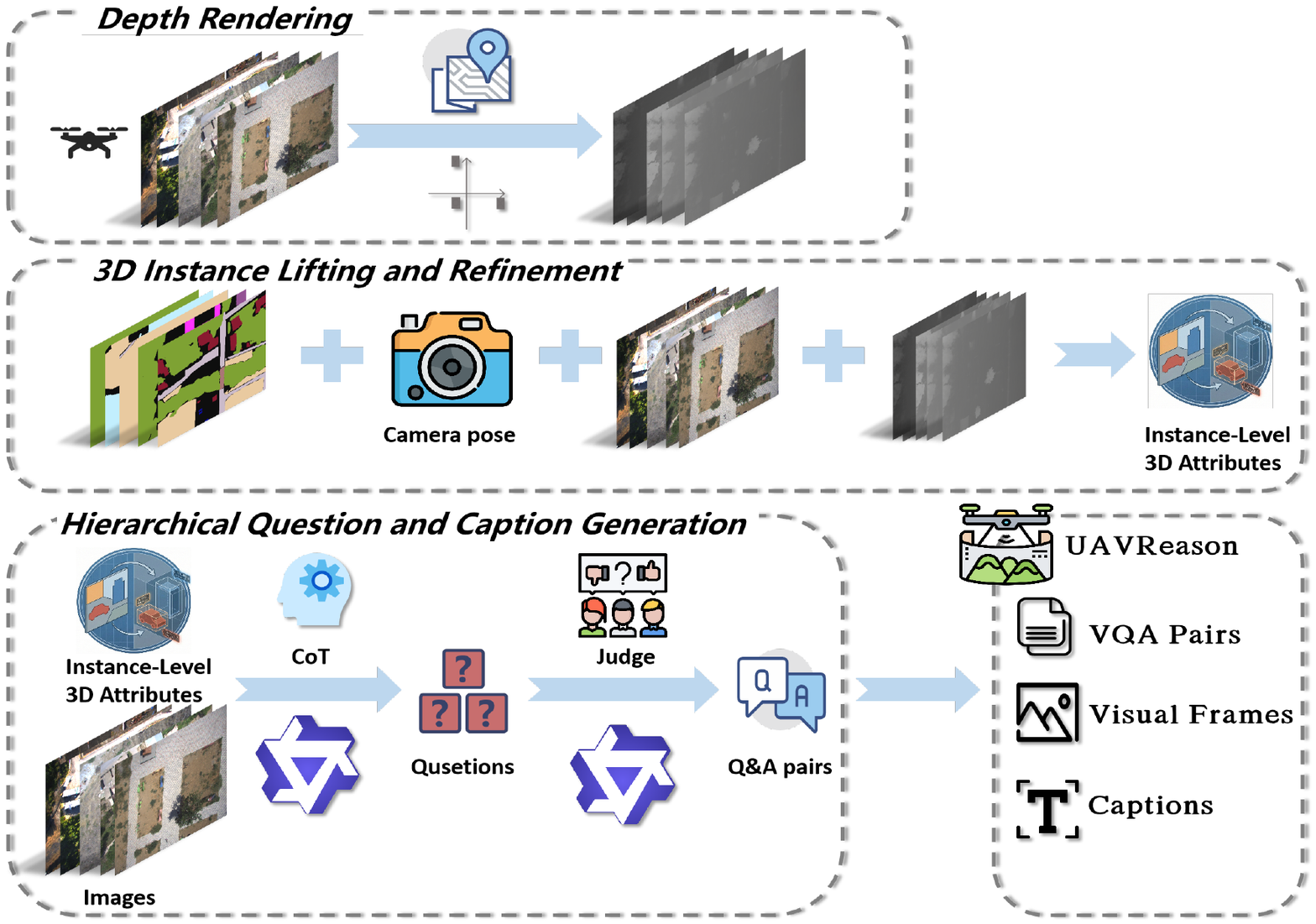}
  \vspace{-.1in}
  \caption{\textbf{UAVReason construction pipeline.}
    We apply the off-the-shelf tools to generate dense, physically grounded annotations for the nadir-view UAV imagery, including depth rendering, 3D instance lifting and refinement, and hierarchical question and caption generation.
    }
  \label{fig:pipeline}
\end{figure}

Different from existing benchmarks in Table~\ref{tab:dataset_comparison_final}, UAVReason is constructed as a unified benchmark for spatio-temporal reasoning and cross-modal generation in nadir-view UAV scenarios (see Figure~\ref{fig:all_task}). To support such capabilities, the benchmark is designed to provide not only semantic annotations, but also precise, metric-level geometric grounding across space and time. As shown in Figure~\ref{fig:pipeline}, we develop an automated pipeline that systematically transforms raw multi-sensor UAV data into geometry-aware representations suitable for reasoning and generation. In particular, our construction leverages UAVScenes~\cite{uavscenes2025} as a high-fidelity data source, which offers synchronized camera and LiDAR streams with accurate sensor calibration across diverse real-world environments, including towns, valleys, and airports. These properties enable dense geometric supervision to be derived directly from the underlying 3D environment, without reliance on manual semantic annotation, while preserving physical consistency.

\textbf{Depth Rendering.} We first obtain dense, pixel-aligned depth as the geometric foundation for subsequent reasoning and question generation. However, raw LiDAR point clouds are inherently sparse in the image plane. To address this, we render dense depth maps from the high-fidelity 3D environment meshes registered to the sensor trajectories in UAVScenes. We configure an off-screen renderer with the exact camera intrinsics and extrinsic pose for each frame. The renderer's Z-buffer provides accurate ground-truth depth supervision.

\begin{figure}[t!]
  \centering
  \vspace{-.1in}
  \includegraphics[width=0.7\linewidth]{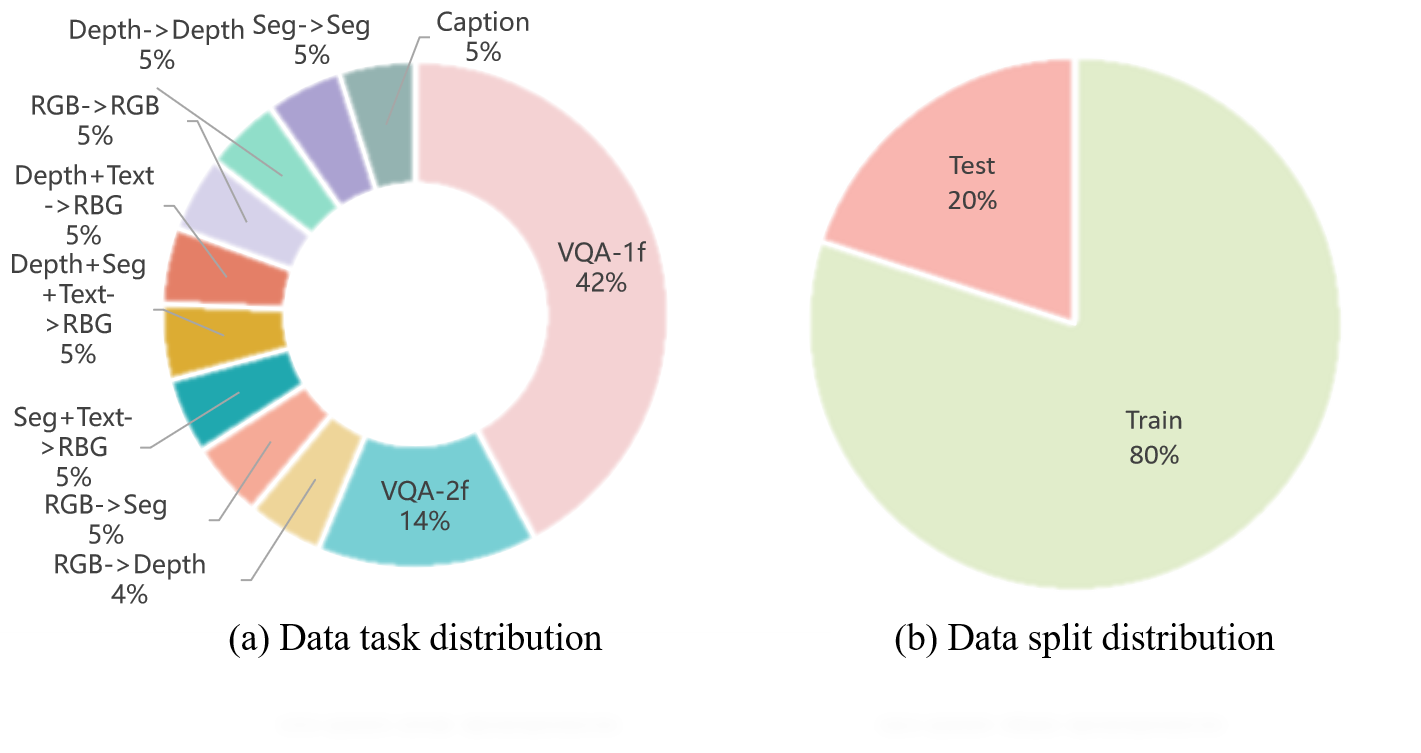}
  \vspace{-.3in} 
  \caption{Task and Data Distribution in UAVReason.}
  \label{fig:data}
  \vspace{-.25in}
\end{figure}

\textbf{3D Instance Lifting and Refinement.}
With dense depth maps available, we derive explicit object-level ground truth to define instance identity and 3D localization, enabling geometry-grounded question--answer pairs on object identity, location, and quantitative spatial relations.
We first extract candidate object regions via connected-component analysis on semantic segmentation masks. In nadir-view imagery, occlusions and limited resolution often merge distinct objects in the image plane. To recover accurate instance boundaries, we leverage depth discontinuities to split connected regions into spatially coherent components, followed by refinement to suppress noisy detections. Specifically, we employ a hybrid non-maximum suppression with a tightness criterion to eliminate erroneous enclosures of smaller instances within larger regions.
Each refined 2D instance mask is then lifted to 3D by back-projecting its pixels into the world coordinate system using calibrated camera intrinsics and rendered depth maps. From the resulting 3D point set, we fit an oriented bounding box (OBB) and compute metric attributes, including the 3D centroid and orientation.
These geometry-grounded object representations serve as the foundation for QA pairs involving object counting, distance estimation, comparative spatial relations, and other quantitative reasoning tasks.

\textbf{Hierarchical Question and Caption Generation.}
Building on the geometry-grounded 3D object instances, we construct questions and captions via a three-step pipeline: \textbf{(1) Contextual Enrichment.} For each frame, we first generate a global scene description to establish overall context. At the instance level, we retrieve object attributes and spatial relations directly from the associated 3D representations, ensuring all references remain faithful to the underlying scene geometry.  \textbf{(2) Taxonomy-Guided Question Generation.} To promote diverse reasoning demands, we create questions spanning multiple levels: basic object perception, metric spatial reasoning, and multi-step logical inference. By leveraging persistent 3D instance identities across frames, we further generate temporal questions that probe object motion, trajectory, and state evolution over time. 
\textbf{(3) Programmatic Answer Construction.}
For each question, the answer is obtained through explicit geometric computation on the 3D ground truth. A \textit{Planner} parses the question into a symbolic execution program (\eg, \texttt{calc\_distance}, \texttt{compare\_count}), which is evaluated by a deterministic \textit{Executor} operating directly on the 3D ground truth. Finally, a \textit{Writer} renders the computed result in natural language. This programmatic approach guarantees metric accuracy, reproducibility, and consistent supervision for spatial and temporal reasoning evaluation.

\begin{figure*}[t!]
  \centering
  \vspace{-.3in}
  \includegraphics[width=\linewidth]{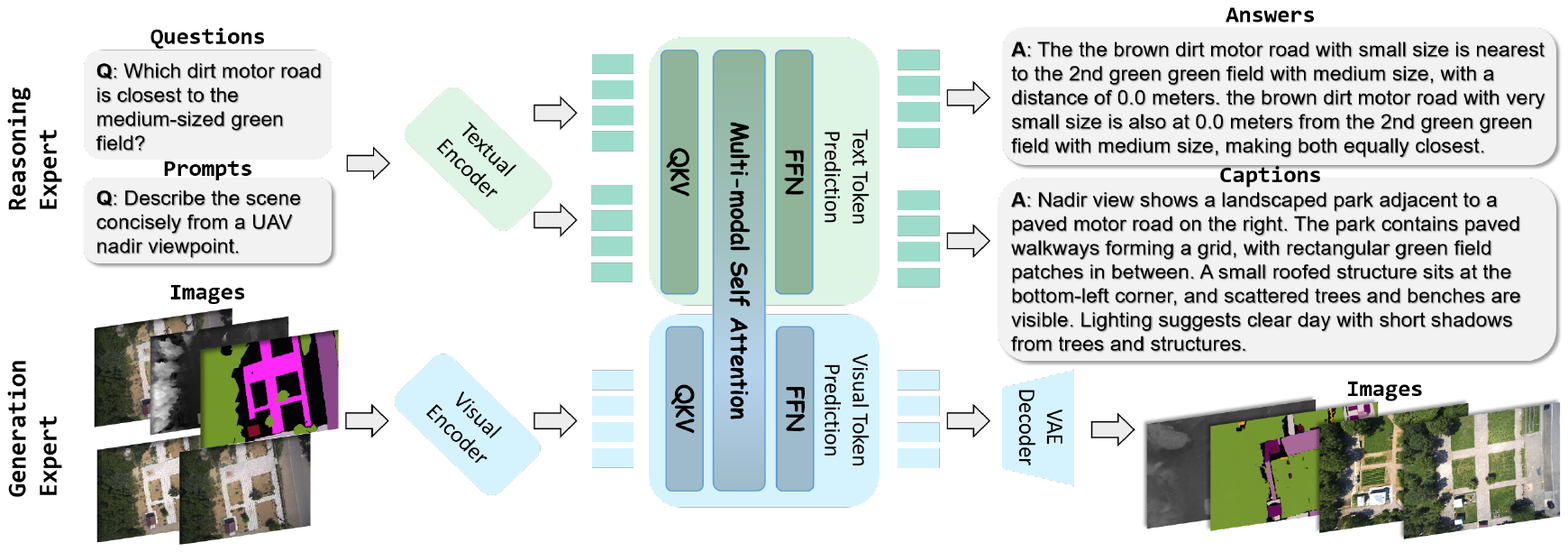}
  \vspace{-5.3cm} 
  \caption{\textbf{Unified multi-task baseline for UAVReason.} We adopt a shared multi-modal transformer backbone with two task-specialized experts: a \textbf{Reasoning Expert} (top) optimized for next-token prediction over language outputs, and a \textbf{Generation Expert} (bottom) optimized for velocity prediction under diffusion-style latent denoising.
Text is tokenized and fused with visual features via multi-modal self-attention, while expert-specific QKV projections and FFNs specialize computation for reasoning versus generation.
The reasoning branch consumes UAV RGB frames with questions (and captioning prompts) to produce textual answers, whereas the generation branch conditions on RGB/depth/segmentation (and optional text) to synthesize target modalities.
This design enables a single backbone to jointly learn language reasoning and pixel-level generation under a consistent UAV domain.}
  \label{fig:arch}
  \vspace{-.1in}
\end{figure*}
\subsection{Task Description}
\label{sec:dataset_description}
\textbf{Task Formulation.}
Our benchmark includes two complementary tracks that share  aligned RGB, depth, and segmentation signals.
As shown Figure~\ref{fig:data}, it contains 23,600 unique frames~(19,901 training and 3,699 testing samples) on 20 distinct maps covering diverse terrains.

\textbf{Reasoning Track.}
To probe semantic grounding under the challenging nadir viewpoint, we provide rich language supervision organized into three core evaluation settings.
UAVReason contains \textbf{273,000} VQA pairs, comprising:
(i) \textbf{204,777} single-frame questions targeting static spatial and compositional reasoning (\textbf{VQA-1F});
and (ii) \textbf{68,223} two-frame questions designed for temporal dynamics.
The latter is further stratified into standard temporal reasoning (\textbf{VQA-2F}) and harder navigational reasoning requiring implicit heading cues (\textbf{VQA-2F$_\text{ihead}$}).
In addition, each of the \textbf{23,600} frames is annotated with a detailed scene-level caption that summarizes global context and salient object relations.
The VQA taxonomy spans \textbf{22} sub-categories grouped into 4 major axes, such as \emph{Object Counting}, \emph{Spatial Reasoning}, \emph{Spatial Comparison}, and \emph{Temporal Reasoning}, enabling fine-grained diagnosis of diverse reasoning ability.

\textbf{Generation Track.}
To enforce pixel-level grounding, each RGB frame is paired with precisely aligned dense depth maps and semantic segmentation masks, supporting eight multimodal generation tasks with a total of \textbf{188,800} samples. These tasks are grouped into three families.
First, \emph{cross-modal translation} (RGB$\to$Depth and RGB$\to$Segmentation) requires recovering geometric and semantic structures from visual appearance alone.
Second, \emph{conditional synthesis} (Text+Depth$\to$RGB, Text+Segmentation$\to$RGB, and Text+Depth+Segmentation$\to$RGB) assesses the generation of photorealistic UAV imagery consistent with both geometric constraints and linguistic intent.
Third, \emph{reconstruction} performs self-supervised reconstruction of RGB, depth, and segmentation, serving as an auxiliary signal to enforce consistency between high-level representations and pixel-level geometric and appearance details.
Together, these families enable unified evaluation of semantic reasoning and pixel-level generation under a consistent UAV data distribution and shared geometric constraints.


\begin{table*}[t]
\centering
\small 
\setlength{\tabcolsep}{1pt} 
\renewcommand{\arraystretch}{1.2}
\caption{\textbf{Results on UAVReason for spatio-temporal VQA and scene captioning.}
We evaluate 3 reasoning settings: single-frame VQA (VQA-1F), two-frame VQA (VQA-2F), and two-frame VQA with implicit heading cues (VQA-2F$_\text{ihead}$), alongside scene captioning.
For VQA, we report Exact Match (EM), token-level F1, and an LLM-based judge score (LLM-J) to measure semantic correctness beyond lexical overlap.
Methods are grouped into general-purpose VLMs, UAV/remote-sensing VLMs, unified generation models, and our unified framework.
\textbf{UAVReason-Bagel} achieves state-of-the-art performance across all tasks. Compared with the ablation row (\textit{Reasoning only}), joint training with generation improves semantic consistency in temporal reasoning (higher LLM-J).}
\begin{tabular}{l ccc ccc ccc ccc}
\toprule

\multirow{2.5}{*}{\textbf{Method}} & 
\multicolumn{3}{c}{\textbf{VQA-1F}} & 
\multicolumn{3}{c}{\textbf{VQA-2F}} & 
\multicolumn{3}{c}{\textbf{VQA-2F$_\text{ihead}$}} & 
\multicolumn{3}{c}{\textbf{Caption}} \\

\cmidrule(lr){2-4} \cmidrule(lr){5-7} \cmidrule(lr){8-10} \cmidrule(lr){11-13}

& EM $\uparrow$ & F1 $\uparrow$ & \textbf{LLM-J} $\uparrow$ & 
EM $\uparrow$ & F1 $\uparrow$ & \textbf{LLM-J} $\uparrow$ & 
EM $\uparrow$ & F1 $\uparrow$ & \textbf{LLM-J} $\uparrow$ & 
B4 $\uparrow$ & C $\uparrow$ & \textbf{LLM-J} $\uparrow$ \\

\midrule

\rowcolor{headergray} 
\multicolumn{13}{l}{\textit{Group A: Off-the-shelf General VLMs / MLLMs}} \\
Qwen2.5-VL-7B & 0.041 & 0.217 & 0.355 & 0.000 & 0.138 & 0.454 & 0.000 & 0.088 & 0.000 & 0.015 & 0.321 & 7.02 \\
Qwen2.5-VL-3B & 0.052 & 0.163 & 0.325 & 0.000 & 0.236 & 0.323 & 0.000 & 0.086 & 0.050 & 0.019 & 0.319 & 6.33 \\
Qwen3-VL-8B & 0.031 & 0.113 & 0.325 & 0.000 & 0.207 & 0.475 & 0.000 & 0.088 & 0.080 & 0.435 & 0.041 & 7.01 \\
Qwen3-VL-4B & 0.031 & 0.132 & 0.340 & 0.000 & 0.279 & 0.414 & 0.000 & 0.041 & 0.070 & 0.037 & 0.513 & 6.99 \\
LLaVA-OneVis-7B & 0.031 & 0.109 & 0.247 & 0.000 & 0.099 & 0.126 & 0.000 & 0.030 & 0.060 & 0.007 & 0.221 & 5.89 \\
InternVL3.5-1B & 0.000 & 0.205 & 0.299 & 0.000 & 0.244 & 0.333 & 0.000 & 0.027 & 0.060 & 0.027 & 0.395 & 6.57 \\
InternVL3.5-2B & 0.062 & 0.163 & 0.294 & 0.000 & 0.140 & 0.414 & 0.140 & 0.865 & 0.180 & 0.024 & 0.416 & 7.03 \\
InternVL3.5-4B & 0.021 & 0.329 & 0.340 & 0.000 & 0.151 & 0.439 & 0.000 & 0.038 & 0.070 & 0.017 & 0.393 & 6.67 \\
InternVL3.5-8B & 0.010 & 0.215 & 0.314 & 0.000 & 0.069 & 0.409 & 0.000 & 0.042 & 0.130 & 0.022 & 0.384 & 6.82 \\

\midrule

\rowcolor{headergray} 
\multicolumn{13}{l}{\textit{Group B: UAV / Remote-Sensing VLMs}} \\
GeoChat-7B & 0.031 & 0.371 & 0.175 & 0.000 & 0.281 & 0.505 & 0.000 & 0.050 & 0.060 & 0.004 & 0.162 & 4.34 \\
RS-LLaVA-1.5-7B & 0.021 & 0.232 & 0.263 & 0.000 & 0.309 & 0.374 & 0.000 & 0.334 & 0.040 & 0.000 & 0.192 & 6.37 \\

\midrule

\rowcolor{headergray} 
\multicolumn{13}{l}{\textit{Group C: Unified Generation Models (Off-the-shelf)}} \\
Janus-Pro-7B & 0.021 & 0.141 & 0.294 & 0.000 & 0.423 & 0.253 & 0.000 & 0.215 & 0.090 & 0.021 & 0.327 & 6.77 \\
Janus-1.3B & 0.010 & 0.354 & 0.160 & 0.000 & 0.532 & 0.444 & 0.000 & 0.448 & 0.090 & 0.012 & 0.330 & 6.01 \\
OmniGen2 & 0.000 & 0.208 & 0.299 & 0.000 & 0.425 & 0.298 & 0.000 & 0.080 & 0.050 & 0.014 & 0.294 & 6.01 \\
Show-o2-7B & 0.052 & 0.106 & 0.286 & 0.000 & 0.081 & 0.455 & 0.000 & 0.078 & 0.340 & 0.011 & 0.179 & 4.95 \\

\midrule

\rowcolor{headergray} 
\multicolumn{13}{l}{\textit{Group D: Our Unified Framework}} \\
Bagel (Pretrain) & 0.000 & 0.394 & 0.28 & 0.000 & 0.427 & 0.250 & 0.000 & 0.798 & 0.090 & 0.022 & 0.387 & 6.68 \\

\rowcolor{highlightrow} 
\textbf{UAVReason-Bagel} & 
\textbf{0.196} & 0.711 & 0.397 & 
0.186 & 0.822 & \textbf{0.600} & 
0.657 & 0.973 & 0.657 & 
0.216 & 1.530 & \textbf{7.69} \\

\color{textgray} \textit{\hspace{1em} Reasoning only} & 
\color{textgray} 0.165 & \color{textgray} \textbf{0.734} & \color{textgray} \textbf{0.454} & 
\color{textgray} \textbf{0.234} & \color{textgray} \textbf{0.855} & \color{textgray} 0.571 & 
\color{textgray} \textbf{0.770} & \color{textgray} \textbf{0.982} & \color{textgray} \textbf{0.770} & 
\color{textgray} \textbf{0.242} & \color{textgray} \textbf{1.741} & \color{textgray} 7.68 \\

\color{textgray} \textit{\hspace{1em} Generation only} & 
\color{textgray} 0.052 & \color{textgray} 0.422 & \color{textgray} 0.330 & 
\color{textgray} 0.000 & \color{textgray} 0.401 & \color{textgray} 0.240 & 
\color{textgray} 0.000 & \color{textgray} 0.801 & \color{textgray} 0.145 & 
\color{textgray} 0.017 & \color{textgray} 0.371 & \color{textgray} 6.63 \\

\bottomrule
\end{tabular}
\label{tab:main_vqa_caption}
\end{table*}
\begin{table*}[t]
\centering
\vspace{-.3in}
\small
\setlength{\tabcolsep}{3.3pt} 
\renewcommand{\arraystretch}{1.2}
\caption{\textbf{Dense perception and conditional cross-modal generation on UAVReason.}
We report 18-class mIoU for semantic segmentation (\textbf{Perc.}).
For conditional image generation, we evaluate three settings: depth+segmentation+text to RGB (dSeg+T$\to$RGB), segmentation+text to RGB (Seg+T$\to$RGB), and depth+text to RGB (d+T$\to$RGB).
Quality is measured by KID ($\downarrow$), CLIP score ($\uparrow$), and DINO similarity ($\uparrow$).
\textbf{UAVReason-Bagel} significantly outperforms off-the-shelf unified generators. The \textit{w/o Understanding} ablation shows that reasoning tasks contribute to better generation quality (lower KID, higher DINO) in sparse-conditioning settings (d+T$\to$RGB).}
\begin{tabular}{l c ccc ccc ccc}
\toprule

\multirow{2.5}{*}{\textbf{Method}} & 
\multicolumn{1}{c}{\textbf{Perc.}} & 
\multicolumn{9}{c}{\textbf{Conditional Generation Quality}} \\
\cmidrule(lr){2-2} \cmidrule(lr){3-11}

& \textbf{Seg} & 
\multicolumn{3}{c}{\textbf{dSeg+T$\to$RGB}} & 
\multicolumn{3}{c}{\textbf{Seg+T$\to$RGB}} & 
\multicolumn{3}{c}{\textbf{d+T$\to$RGB}} \\
\cmidrule(lr){3-5} \cmidrule(lr){6-8} \cmidrule(lr){9-11}

& mIoU$\uparrow$ & 
KID$\downarrow$ & CLIP$\uparrow$ & DINO$\uparrow$ & 
KID$\downarrow$ & CLIP$\uparrow$ & DINO$\uparrow$ & 
KID$\downarrow$ & CLIP$\uparrow$ & DINO$\uparrow$ \\

\midrule

\rowcolor{headergray}
\multicolumn{11}{l}{\textit{Off-the-shelf Unified Generators}} \\
OmniGen2 & 0.037 & 
0.094 & 25.46 & 0.317 & 
0.093 & 24.29 & 0.507 & 
0.094 & 24.40 & 0.378 \\
Show-o2-7B & 0.023 & 
-- & -- & -- & 
0.109 & 25.67 & \textbf{0.537} & 
0.118 & 23.81 & 0.507 \\

\midrule

\rowcolor{headergray}
\multicolumn{11}{l}{\textit{Our Unified Framework}} \\
Bagel (Pretrain) & 0.033 & 
0.078 & 25.69 & 0.513 & 
0.083 & 26.97 & 0.509 & 
0.082 & 26.38 & 0.565 \\

\rowcolor{highlightrow}
\textbf{UAVReason-Bagel} & 
\textbf{0.143} & 
\textbf{0.048} & \textbf{29.52} & 0.648 & 
\textbf{0.050} & \textbf{29.41} & 0.524 & 
\textbf{0.036} & \textbf{29.23} & \textbf{0.609} \\

\color{textgray} \textit{\hspace{1em} Reasoning only} & 
\color{textgray} 0.044 & 
\color{textgray} 0.082 & \color{textgray} 25.41 & \color{textgray} 0.513 & 
\color{textgray} 0.080 & \color{textgray} 26.74 & \color{textgray} 0.522 & 
\color{textgray} 0.084 & \color{textgray} 25.76 & \color{textgray} 0.565 \\

\color{textgray} \textit{\hspace{1em} Generation only} & 
\color{textgray} 0.143 & 
\color{textgray} 0.051 & \color{textgray} 28.59 & \color{textgray} \textbf{0.657} & 
\color{textgray} 0.050 & \color{textgray} 29.01 & \color{textgray} 0.536 & 
\color{textgray} 0.052 & \color{textgray} 28.86 & \color{textgray} 0.594 \\
\bottomrule
\end{tabular}
\label{tab:main_gen_integrated}
\vspace{-.15in}
\end{table*}

\vspace{-.1in}
\section{Method} 
\label{sec:method}
\vspace{-.05in}
In this work, rather than pursuing the strongest possible backbone, we focus on quantifying relative gains from jointly training complementary reasoning and generation tasks. To this end, we introduce \textbf{UAVReason-Bagel} (Figure~\ref{fig:arch}), a specialized adaptation of the unified multimodal foundation model BAGEL~\citep{BAGEL}. 
Building upon BAGEL's unified understanding-and-generation paradigm, UAVReason-Bagel employs a shared multimodal transformer backbone while decoupling task-specific computation through a Mixture-of-Transformer-Experts (MoT) design. In particular, a dedicated Generation Expert is optimized for velocity prediction over continuous visual latents within a Rectified Flow framework, while a dedicated Understanding Expert is optimized for discrete next-token prediction over language outputs. This architectural separation alleviates gradient conflicts arising from jointly optimizing discrete linguistic objectives and continuous visual generation objectives, thereby enabling a single backbone to effectively acquire both high-level spatio-temporal reasoning and pixel-level grounding in the UAV domain.

\textbf{Unified Architecture.}
We follow a Mixture-of-Transformer-Experts (MoT) architecture and unify all UAVReason tasks under a single sequence-to-sequence formulation.
For textual inputs (\eg, questions, instructions, or prompts), the text encoder produces corresponding token sequences. For visual inputs (\eg, RGB images, depth maps, segmentation masks, or temporal sequences), the visual encoder transforms them into visual representations, which are subsequently fused with text tokens.
An autoregressive LLM decoder then generates the target sequence autoregressively. Depending on the task, it produces either text tokens for reasoning-oriented tasks or visual predictions for generation tasks. In the latter case, the predicted visual representations are decoded into images or dense geometric maps via a VAE decoder. This unified backbone enables language-level reasoning and pixel-level generation to share representations and supervisory signals within a single model, fostering mutual reinforcement across modalities and task families.

\textbf{Multi-Task Learning Strategy.}
We train UAVReason-Bagel by jointly optimizing reasoning  and generation objectives. The total training loss is formulated as:
\begin{equation}
    \mathcal{L}_{\text{total}} = \lambda_{\text{CE}} \, \mathcal{L}_{\text{CE}} + \lambda_{\text{MSE}} \, \mathcal{L}_{\text{MSE}},
\label{eq:loss_total}
\end{equation}
where $\mathcal{L}_{\text{CE}}$ is the cross-entropy loss applied to language generation and discrete segmentation prediction, and $\mathcal{L}_{\text{MSE}}$ is the mean squared error supervising continuous depth estimation and pixel-level reconstruction.
To counteract the \emph{semantic floating} phenomenon prevalent in nadir-view UAV imagery, where objects often lack strong contextual cues due to limited viewpoint variation, we deliberately emphasize geometry-aware supervision. This is achieved by setting $\lambda_{\text{MSE}} : \lambda_{\text{CE}} = 2:1$ and adopting a 2:1 data sampling ratio favoring generation tasks over reasoning tasks.
Intuitively, stronger pixel-level geometric supervision encourages the model to internalize robust structural priors (\eg, depth discontinuities, relative scaling, and layout consistency) directly from dense ground-truth signals. These priors, in turn, anchor high-level semantic reasoning to physically plausible configurations, reducing hallucinations and promoting spatially grounded interpretations under the nadir viewpoint
\vspace{-.05in}
\section{Experiment}
\label{sec:experiment}
\vspace{-.05in}
\subsection{Spatio-temporal VQA and Captioning} 
We compare UAVReason-Bagel against a broad spectrum of baselines, including (i) off-the-shelf general VLMs trained primarily on ground-centric data, (ii) specialized UAV/remote-sensing models, and (iii) unified understanding-generation models. Table~\ref{tab:main_vqa_caption} reports results across three VQA settings (VQA-1F, VQA-2F, VQA-2F$_\text{ihead}$) and scene captioning. We find that the nadir-view domain shift challenges general-purpose VLMs.
As shown in Group A, most general VLMs obtain near-zero EM across VQA-1F and VQA-2F, suggesting that producing \emph{exact}, physically grounded answers remains difficult under tiny object scales and ambiguous orientations. While token-level F1 can be non-trivial for some models (\eg, InternVL3.5-4B reaches 0.329 on VQA-1F), the corresponding LLM-J scores are typically modest (about 0.25--0.35), indicating that partial lexical overlap does not reliably translate to semantically correct spatial reasoning.
This gap becomes more evident in the orientation-sensitive VQA-2F$_\text{ihead}$ setting, where LLM-J remains low for most general VLMs, reflecting limited ability to leverage heading cues for distinguishing relative spatial relations. In contrast, unified fine-tuning substantially improves spatio-temporal reasoning.
Specifically, UAVReason-Bagel (Group D) achieves consistent gains over the pre-trained baseline. Fine-tuning improves VQA-1F from 0.394 to 0.711 in F1 (EM from 0.000 to 0.196), and yields stronger temporal reasoning on VQA-2F (0.822 F1 and 0.600 LLM-J versus 0.427 F1 and 0.250 LLM-J for Bagel (Pretrain)).
Notably, with implicit heading cues (VQA-2F$_\text{ihead}$), UAVReason-Bagel reaches 0.973 F1 and 0.657 LLM-J, 
suggesting that the model can effectively leverage orientation signals to reason about relative spatial relations in aerial scenes.

For captioning, content relevance matters beyond n-gram fluency. For scene captioning, UAVReason-Bagel achieves the best overall performance with CIDEr 1.530 and the highest LLM-J 7.69. We also observe that some general VLMs attain relatively high BLEU-4 scores, while exhibiting much lower CIDEr scores (\eg, Qwen3-VL-8B). This highlights that n-gram fluency alone is insufficient for UAV scenes, where accurately identifying salient objects and their spatial relations is critical. 
CIDEr and LLM-J better reflect content relevance and semantic correctness. Beyond captioning quality, we further observe that generation objectives improve semantic robustness in temporal reasoning.
The \textit{reasoning only} ablation reveals a non-trivial interaction between understanding and generation supervision.
Removing generation objectives slightly increases certain lexical-overlap metrics (\eg, VQA-1F F1 0.734 vs.\ 0.711), but reduces semantic correctness on temporal reasoning under LLM-J (VQA-2F LLM-J drops from 0.600 to 0.571).
This suggests that dense generation losses encourage learning geometry-aware structural priors, which in turn improve semantically consistent reasoning in challenging spatio-temporal settings.
\vspace{-.05in}
\subsection{Dense Perception and Generation}
\label{subsec:exp_dense_gen}
\vspace{-.05in}
As shown in Table~\ref{tab:main_gen_integrated}, we summarize performance on dense perception (RGB$\to$Seg) and three conditional generation settings. We observe that off-the-shelf unified generators struggle with aerial dense grounding. Models trained on broad internet distributions exhibit limited capability in UAV dense prediction.
OmniGen2~\cite{omnigen2} and Show-o2-7B~\cite{show-o} obtain very low segmentation performance (mIoU 0.037 and 0.023), and even the pretrained Bagel baseline remains low (0.033), indicating that nadir-view dense grounding requires domain-specific supervision. In contrast, UAVReason-Bagel improves dense perception and geometry-conditioned synthesis. With unified multi-task fine-tuning, UAVReason-Bagel achieves 0.143 mIoU on segmentation, substantially outperforming all baselines.
This improved geometric grounding ability further translates to better conditional synthesis quality.
For dSeg+T$\to$RGB, UAVReason-Bagel reduces KID from 0.078 to 0.048 while increasing CLIP score from 25.69 to 29.52 and DINO score from 0.513 to 0.648, indicating better realism, text alignment, and structural fidelity with the corresponding ground truth.
Similar improvements are observed for Seg+T$\to$RGB and d+T$\to$RGB, especially in CLIP scores, suggesting stronger controllability under UAV prompts.

We further observe that supervision from understanding tasks provides an implicit regularization effect on generation. To examine this effect, we perform a \textit{generation-only} ablation by removing all language-based understanding objectives (\ie, VQA and captioning) during training, while keeping the generation losses unchanged.
Although segmentation performance remains stable (mIoU = 0.143), generation quality degrades significantly under depth-only conditioning (d+T$\rightarrow$RGB): KID increases from 0.036 to 0.052, and DINO similarity drops from 0.609 to 0.594.
These results indicate that language-level understanding objectives regularize the shared representation space, enabling the model to infer correct semantics and realistic textures even when visual inputs are sparse or ambiguous. \textbf{Please refer to the \hyperref[sec:appendix_start]{Appendix} for more generalization results and ablation studies.}
\vspace{-.05in}
\section{Conclusion}
\label{sec:conclusion}
\vspace{-.05in}
Standard VLMs struggle to generalize to the aerial domain, primarily due to \emph{insufficient spatial grounding} under the ambiguous nadir viewpoint.
To bridge this gap, we introduce \textbf{UAVReason}, a unified benchmark that jointly evaluates spatio-temporal reasoning and pixel-level generation in UAV scenes.
Our baseline model, \textbf{UAVReason-Bagel}, shows that dense geometric supervision acts as a robust structural prior.
Empirical results confirm our hypothesis that \emph{``Synthesis promotes Analysis''}: generative objectives significantly enhance semantic robustness in temporal reasoning while maintaining geometry-consistent synthesis.
We hope this work paves the way for future research on physically grounded embodied aerial agents.

\section{Limitations}
\label{sec:limitations}

UAVReason has three main limitations. First, although it is built from high-fidelity UAV data with geometry-grounded annotations, it may not cover all real-world conditions, such as adverse weather, sensor degradation, extreme altitudes, or highly dynamic scenes. Second, the automatic construction pipeline may still introduce errors in instance refinement, caption generation, or question construction. Third, UAVReason-Bagel uses a 7B-parameter unified backbone with full-parameter fine-tuning, which requires substantial GPU resources and is not optimized for real-time edge deployment.

{
\small
\bibliographystyle{plainnat}
\bibliography{ref}
}






\clearpage
\appendix

\phantomsection
\label{sec:appendix_start}

\begin{center}
    {\LARGE\bfseries
    UAVReason: A Unified, Large-Scale Benchmark for Multimodal Aerial Scene Reasoning and Generation
    \par}

    \vspace{0.8em}

    {\Large\itshape Supplementary Materials\par}
\end{center}

\vspace{1.2em}

\noindent{\Large\bfseries Contents}

\vspace{1.2em}

\appitem{sec:impl}{A \quad Implementation Details}
\appitem{sec:eval}{B \quad Evaluation Protocol}
\appitem{sec:ablation}{C \quad Ablation Studies}
\appitem{sec:appendix_qualitative}{D \quad More Qualitative Analysis}
\appsubitem{sec:qual:vqa}{D.1 \quad Spatio-temporal VQA and Captioning Qualitative Analysis}
\appsubitem{sec:qual:gen}{D.2 \quad Dense Perception and Generation Qualitative Analysis}

\vspace{0.55em}
\appitem{sec:appendix_prompts}{E \quad Prompt Engineering Details}

\hrule height 1pt
\vspace{2pt}
\hrule height 0.4pt

\section{Implementation Details}
\label{sec:impl}
\paragraph{Optimization Setup.}
We perform full-parameter fine-tuning of the Bagel-7B-MoT backbone on 8 GPUs for 30,000 steps.
We use AdamW with a peak learning rate of $8\times 10^{-6}$ and a 1,000-step linear warmup, followed by a cosine decay schedule.
To test our hypothesis that geometry supervision provides a structural constraints for aerial reasoning, we enforce the loss weighting in Eq.~\ref{eq:loss_total} as
$\lambda_{\text{MSE}}:\lambda_{\text{CE}}=\mathbf{2:1}$.
This design targets the \emph{semantic floating} issue by prioritizing pixel-level fidelity over language-only objectives, which helps reduce the tendency to overfit to dataset-specific language priors and encourages grounded adaptation to the nadir UAV domain.

\paragraph{Efficient Input Handling.}
UAVReason involves heterogeneous input patterns, ranging from single-frame spatial queries to two-frame temporal sequences.
To process these mixed-length multimodal samples efficiently, we adopt dynamic sequence packing with a maximum budget of 11,520 tokens per batch.
In particular, since spatio-temporal VQA consumes a pair of frames $(X_{v1}, X_{v2})$ together with question tokens, this strategy ensures high training throughput without truncating the long-context inputs required for temporal reasoning.

\section{Evaluation Protocol}
\label{sec:eval}

We adopt a strict, task-aligned evaluation protocol to handle the open-ended nature of aerial reasoning and to quantify generative fidelity. Beyond naive string matching, we report conventional automatic metrics together with Large Language Model (LLM) based semantic evaluation. For LLM based evaluation, we use \textbf{Qwen3-VL-235B}~\cite{bai2025qwen3vltechnicalreport} as the judge model and apply fixed prompt templates and rubrics to ensure consistent and reproducible scoring. We denote these semantic scores as \textbf{LLM-Judge} (LLM-J), with task-specific scales used across VQA and captioning.

\paragraph{VQA (Single-frame and Two-frame).}
We report \textbf{Exact Match (EM)} and \textbf{Token-level F1} after standard answer normalization, including lowercasing, punctuation and article removal, and whitespace canonicalization. For numerically grounded questions (\eg, distance, area, speed), we additionally report a tolerance-aware correctness criterion, where a prediction is considered correct if it falls within a relative tolerance $\tau = 5\%$ or an absolute tolerance $\epsilon$ of the ground truth, with $\epsilon$ depending on the physical unit. To assess semantic correctness beyond lexical overlap, we further use an LLM judge, \ie, Qwen3-VL-235B~\cite{bai2025qwen3vltechnicalreport},  that compares the model prediction with the ground-truth answer under two complementary notions of correctness. 
The judge outputs two boolean fields in JSON format: \texttt{semantic\_correct} and \texttt{strict\_correct}. 
The former allows paraphrases and minor numerical deviations as long as key facts and relations are preserved, while the latter requires exact matches for categorical and counting answers.
Based on these two fields, we compute the VQA LLM-J score by mapping \{(\texttt{semantic\_correct}, \texttt{strict\_correct})\} to a three-level score: 1.0 if both are true, 0.5 if only \texttt{semantic\_correct} is true, and 0.0 otherwise, then averaging over the test set.

\begin{table*}[t]
\centering
\small
\setlength{\tabcolsep}{4.5pt}
\renewcommand{\arraystretch}{1.25}

\begin{tabular}{l cc cc cc cc}
\toprule

\multirow{2.5}{*}{\textbf{Resolution}} & 
\multicolumn{2}{c}{\textbf{VQA-1F}} & 
\multicolumn{2}{c}{\textbf{VQA-2F}} & 
\multicolumn{2}{c}{\textbf{VQA-2F$_\text{ihead}$}} & 
\multicolumn{2}{c}{\textbf{Caption}} \\
\cmidrule(lr){2-3} \cmidrule(lr){4-5} \cmidrule(lr){6-7} \cmidrule(lr){8-9}
& EM $\uparrow$ & F1 $\uparrow$ & 
EM $\uparrow$ & F1 $\uparrow$ & 
EM $\uparrow$ & F1 $\uparrow$ & 
B4 $\uparrow$ & CIDEr $\uparrow$ \\

\midrule

256 & 
\textbf{0.196} & 0.711 & 
0.186 & 0.822 & 
0.657 & 0.973 & 
0.216 & 1.530 \\

384 & 
0.186 & \textbf{0.727} & 
0.197 & \textbf{0.850} & 
0.793 & 0.980 & 
\textbf{0.254} & \textbf{1.800} \\

512 & 
0.189 & 0.725 & 
\textbf{0.200} & 0.847 & 
\textbf{0.868} & \textbf{0.990} & 
0.236 & 1.663 \\

\bottomrule
\end{tabular}
\caption{\textbf{Effect of input image resolution on UAVReason understanding tasks.}
Resolution 384 yields the best overall trade-off, achieving the strongest captioning scores and competitive VQA performance. Higher resolution (512) specifically benefits heading-aware VQA, likely due to better visibility of subtle orientation cues.}
\label{tab:ablation_resolution}
\end{table*}

\begin{table*}[t]
\centering
\small 
\setlength{\tabcolsep}{3pt} 
\renewcommand{\arraystretch}{1.25}

\begin{tabular}{l c ccc ccc ccc}
\toprule

\multirow{2.5}{*}{\textbf{Resolution}} & 
\multicolumn{1}{c}{\textbf{Perc.}} & 
\multicolumn{9}{c}{\textbf{Conditional Generation Quality}} \\
\cmidrule(lr){2-2} \cmidrule(lr){3-11}
& \textbf{Seg} & 
\multicolumn{3}{c}{\textbf{dSeg+T$\to$RGB}} & 
\multicolumn{3}{c}{\textbf{Seg+T$\to$RGB}} & 
\multicolumn{3}{c}{\textbf{d+T$\to$RGB}} \\
\cmidrule(lr){3-5} \cmidrule(lr){6-8} \cmidrule(lr){9-11}
& mIoU$\uparrow$ & 
KID$\downarrow$ & CLIP$\uparrow$ & DINO$\uparrow$ & 
KID$\downarrow$ & CLIP$\uparrow$ & DINO$\uparrow$ & 
KID$\downarrow$ & CLIP$\uparrow$ & DINO$\uparrow$ \\

\midrule

256 & 
0.143 & 
0.048 & \textbf{29.52} & 0.648 & 
0.050 & 29.41 & 0.524 & 
0.036 & 29.23 & 0.609 \\

384 & 
\textbf{0.194} & 
0.041 & 29.14 & 0.554 & 
0.032 & \textbf{30.23} & 0.696 & 
0.028 & 29.20 & 0.582 \\

512 & 
0.144 & 
\textbf{0.024} & 28.02 & \textbf{0.697} & 
\textbf{0.024} & 29.52 & \textbf{0.701} & 
\textbf{0.018} & \textbf{29.38} & \textbf{0.724} \\

\bottomrule
\end{tabular}
\caption{\textbf{Effect of input resolution on dense perception and conditional generation.}
Higher resolution consistently improves generation fidelity (lowest KID at 512) and structural consistency (highest DINO at 512). However, segmentation performance peaks at 384.}
\label{tab:ablation_res_gen}
\end{table*}


\paragraph{Scene Captioning.}
We evaluate caption quality using \textbf{BLEU-4} for n-gram fluency and \textbf{CIDEr} for content relevance. CIDEr is particularly important for UAV imagery because it upweights informative geometry-related terms and relations (\eg, \textit{paved road}, \textit{roof}, \textit{intersection}) over generic descriptions. Similarly, we report an LLM based semantic score. The judge is provided with the candidate caption, a set of reference captions, and the image, and assigns an integer score on a 1 to 10 scale. The rubric evaluates semantic similarity to the references, coverage of salient objects and relations, and penalizes hallucinations of objects not supported by the image or references. The final caption LLM-J score is the average judge score across the test set.

\paragraph{Geometry Perception (RGB$\to$Segmentation).}
For semantic segmentation, we report \textbf{mean Intersection over Union (mIoU)} across 18 classes. Predictions are converted from the segmentation color palette to discrete class indices via nearest-neighbor matching prior to evaluation, providing a robust measure of boundary adherence and region accuracy against ground-truth masks.

\paragraph{Conditional Generation.}
Since pixel-wise metrics (\eg, PSNR) often penalize plausible multimodal outputs, we prioritize feature-level and distribution-level metrics. We report \textbf{KID} ($\downarrow$) to measure distribution distance between generated and real UAV images, \textbf{CLIP Score} ($\uparrow$) to assess text-image semantic alignment, and \textbf{DINOv2 Similarity} ($\uparrow$) to quantify structural fidelity to the paired ground truth. Together, these metrics capture complementary aspects of generation quality, including realism, semantic controllability, and geometric consistency.

\section{Ablation Studies}
\label{sec:ablation}

\paragraph{Ablation on Image Resolution.} Resolution trades off efficiency, dense grounding, and heading-aware reasoning.
Tables~\ref{tab:ablation_resolution} and \ref{tab:ablation_res_gen} analyze the effect of input resolution.
For understanding, 384 offers the best overall trade-off, improving captioning (CIDEr 1.800) while remaining competitive on VQA (Table~\ref{tab:ablation_resolution}).
Meanwhile, 512 yields the strongest heading-aware setting (VQA-2F$_\text{ihead}$ with EM 0.868 and F1 0.990), suggesting that higher resolution benefits fine-grained orientation cues.
For generation, higher resolution generally improves distribution and structural fidelity (Table~\ref{tab:ablation_res_gen}), with 512 achieving the lowest KID and highest DINO across conditional synthesis settings.
However, given the quadratic cost increase, we adopt 384 as a practical default for balanced performance, reserving 512 for tasks that prioritize high-fidelity synthesis or heading-aware reasoning.

\paragraph{Ablation on Learning Rate.} 
We investigate the impact of the learning rate ($\lambda$) to find the optimal balance between multimodal understanding and pixel-level generation. 
As shown in Table~\ref{tab:ablation_lr_und} and Table~\ref{tab:ablation_lr_gen}, the choice of $\lambda$ presents a trade-off. 
A lower rate ($5\mathrm{e}{-6}$) tends to favor text-heavy tasks like Captioning and multi-frame VQA, likely because it preserves pre-trained language knowledge better. 
However, for dense prediction tasks, our default setting of $\lambda=8\mathrm{e}{-6}$ proves to be critical. 
Specifically, in Table~\ref{tab:ablation_lr_gen}, $8\mathrm{e}{-6}$ yields the best Segmentation performance (0.143 mIoU) and consistently achieves the lowest KID scores across all generation conditions. 
When $\lambda$ is increased to $1\mathrm{e}{-5}$, we observe a significant drop in segmentation stability (0.093 mIoU). 
Considering that $8\mathrm{e}{-6}$ also maintains the highest exact match score on single-frame VQA (0.196 in Table~\ref{tab:ablation_lr_und}), we select it as the final configuration to ensure robust performance across both understanding and generative capabilities.
\begin{table*}[t]
\centering
\small
\setlength{\tabcolsep}{4.5pt}
\renewcommand{\arraystretch}{1.25}

\begin{tabular}{l cc cc cc cc}
\toprule

\multirow{2.5}{*}{\textbf{Learning Rate}} & 
\multicolumn{2}{c}{\textbf{VQA-1F}} & 
\multicolumn{2}{c}{\textbf{VQA-2F}} & 
\multicolumn{2}{c}{\textbf{VQA-2F$_\text{ihead}$}} & 
\multicolumn{2}{c}{\textbf{Caption}} \\
\cmidrule(lr){2-3} \cmidrule(lr){4-5} \cmidrule(lr){6-7} \cmidrule(lr){8-9}
& EM $\uparrow$ & F1 $\uparrow$ & 
EM $\uparrow$ & F1 $\uparrow$ & 
EM $\uparrow$ & F1 $\uparrow$ & 
B4 $\uparrow$ & CIDEr $\uparrow$ \\

\midrule

8e-6 & 
\textbf{0.196} & 0.711 & 
0.186 & 0.822 & 
0.657 & 0.973 & 
0.216 & 1.530 \\

1e-5 & 
0.186 & \textbf{0.726} & 
0.159 & 0.839 & 
0.725 & 0.725 & 
0.225 & 1.540 \\

5e-6 & 
0.189 & 0.725 & 
\textbf{0.200} & \textbf{0.847} & 
\textbf{0.868} & \textbf{0.990} & 
\textbf{0.236} & \textbf{1.663} \\

\bottomrule
\end{tabular}
\vspace{-.1in}
\caption{\textbf{Impact of learning rate on understanding tasks.} 
Our default setting ($8\mathrm{e}{-6}$) achieves the highest accuracy (EM) on single-frame VQA, which is the foundational capability for the model. While a lower rate ($5\mathrm{e}{-6}$) benefits long-text captioning, $8\mathrm{e}{-6}$ provides a robust baseline across different question types.}
\label{tab:ablation_lr_und}
\end{table*}

\begin{table*}[t]
\centering
\small 
\setlength{\tabcolsep}{3pt} 
\renewcommand{\arraystretch}{1.25}

\begin{tabular}{l c ccc ccc ccc}
\toprule

\multirow{2.5}{*}{\textbf{Learning Rate}} & 
\multicolumn{1}{c}{\textbf{Perc.}} & 
\multicolumn{9}{c}{\textbf{Conditional Generation Quality}} \\
\cmidrule(lr){2-2} \cmidrule(lr){3-11}
& \textbf{Seg} & 
\multicolumn{3}{c}{\textbf{dSeg+T$\to$RGB}} & 
\multicolumn{3}{c}{\textbf{Seg+T$\to$RGB}} & 
\multicolumn{3}{c}{\textbf{d+T$\to$RGB}} \\
\cmidrule(lr){3-5} \cmidrule(lr){6-8} \cmidrule(lr){9-11}
& mIoU$\uparrow$ & 
KID$\downarrow$ & CLIP$\uparrow$ & DINO$\uparrow$ & 
KID$\downarrow$ & CLIP$\uparrow$ & DINO$\uparrow$ & 
KID$\downarrow$ & CLIP$\uparrow$ & DINO$\uparrow$ \\

\midrule

8e-6 & 
\textbf{0.143} & 
\textbf{0.048} & 29.52 & \textbf{0.648} & 
\textbf{0.050} & 29.41 & \textbf{0.524} & 
0.036 & \textbf{29.23} & \textbf{0.609} \\

1e-5 & 
0.093 & 
0.070 & \textbf{29.73} & 0.563 & 
0.073 & 28.91 & 0.443 & 
\textbf{0.028} & 28.23 & 0.486 \\

5e-6 & 
0.135 & 
0.067 & 28.85 & 0.582 & 
0.071 & \textbf{29.49} & 0.514 & 
0.062 & 28.84 & 0.507 \\

\bottomrule
\end{tabular}
\vspace{-.1in}
\caption{\textbf{Effect of learning rate on dense perception and generation.} 
Our selected learning rate ($8\mathrm{e}{-6}$) significantly outperforms others in segmentation (mIoU) and image generation fidelity (KID/DINO), indicating that pixel-level tasks require this specific optimization magnitude to converge effectively.}
\label{tab:ablation_lr_gen}
\end{table*}

\section{More Qualitative Analysis}
\label{sec:appendix_qualitative}

\subsection{Spatio-temporal VQA and Captioning Qualitative Analysis}
\label{sec:qual:vqa}
\begin{figure*}[t!]
  \centering
  \vspace{-3.6cm} 
  \includegraphics[width=\linewidth]{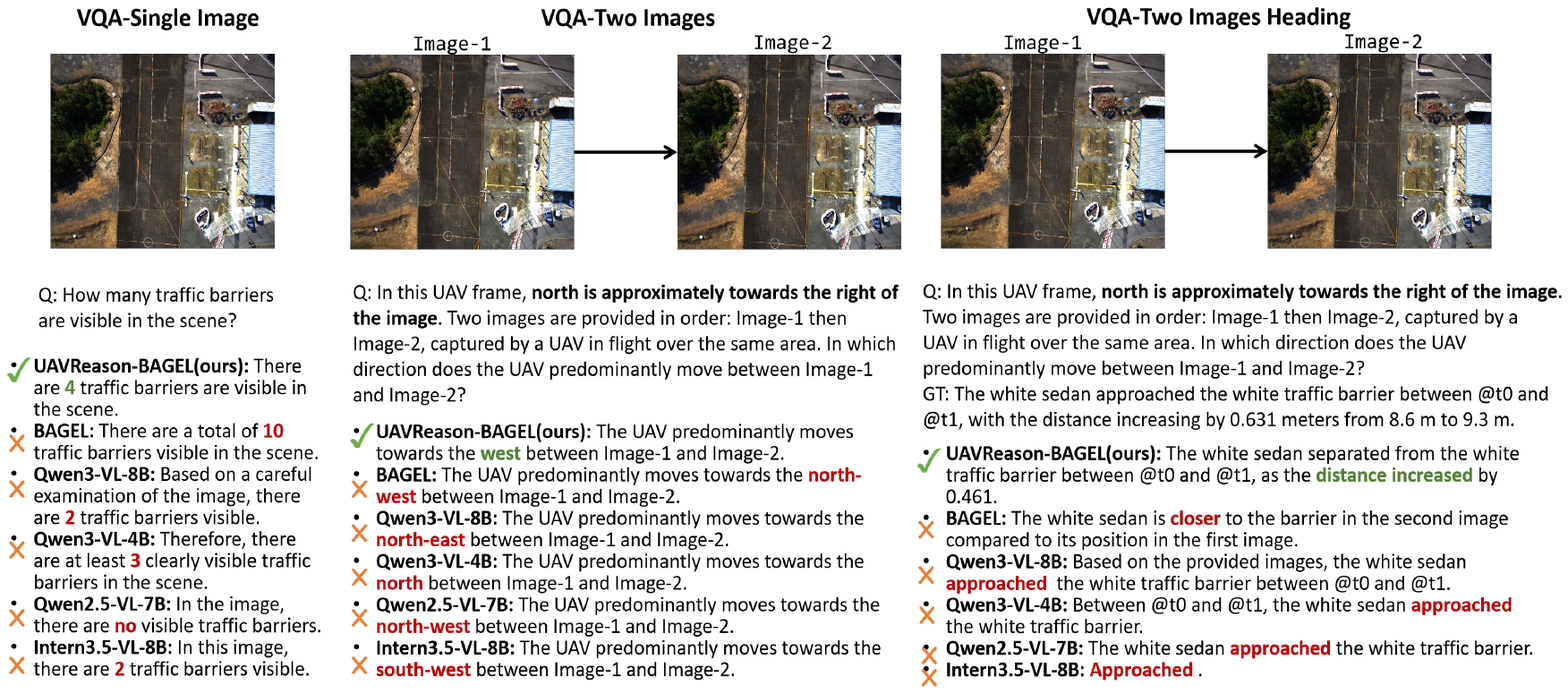}
  \vspace{-1.9cm} 
 \caption{\textbf{Qualitative comparison on fine-grained spatio-temporal reasoning.} 
  We evaluate \textbf{UAVReason-Bagel} (Ours) against leading general-domain VLMs (\eg, Qwen2.5-VL~\cite{qwen2.5-vl}, InternVL~\cite{internvl-2.5}) across three challenging aerial tasks.
  \textbf{Left (Counting):} Our model accurately counts tiny, densely packed instances (traffic barriers), whereas baselines suffer from severe miss-detection or count hallucinations.
  \textbf{Middle (Navigation):} Guided by explicit orientation cues (\eg, ``North is right''), our model correctly aligns the heading to infer the ``West'' trajectory, while general VLMs fail to adapt compass directions to the nadir view.
  \textbf{Right (Temporal Dynamics):} Our model successfully detects fine-grained status changes (separation) and quantifies distance shifts, whereas baselines often hallucinate the opposite motion trend (\eg, ``approaching'').}
  \label{fig:qualitative_reasoning}
   \vspace{-.15in} 
\end{figure*}

\begin{figure*}[t!]
  \centering
  \includegraphics[width=\linewidth]{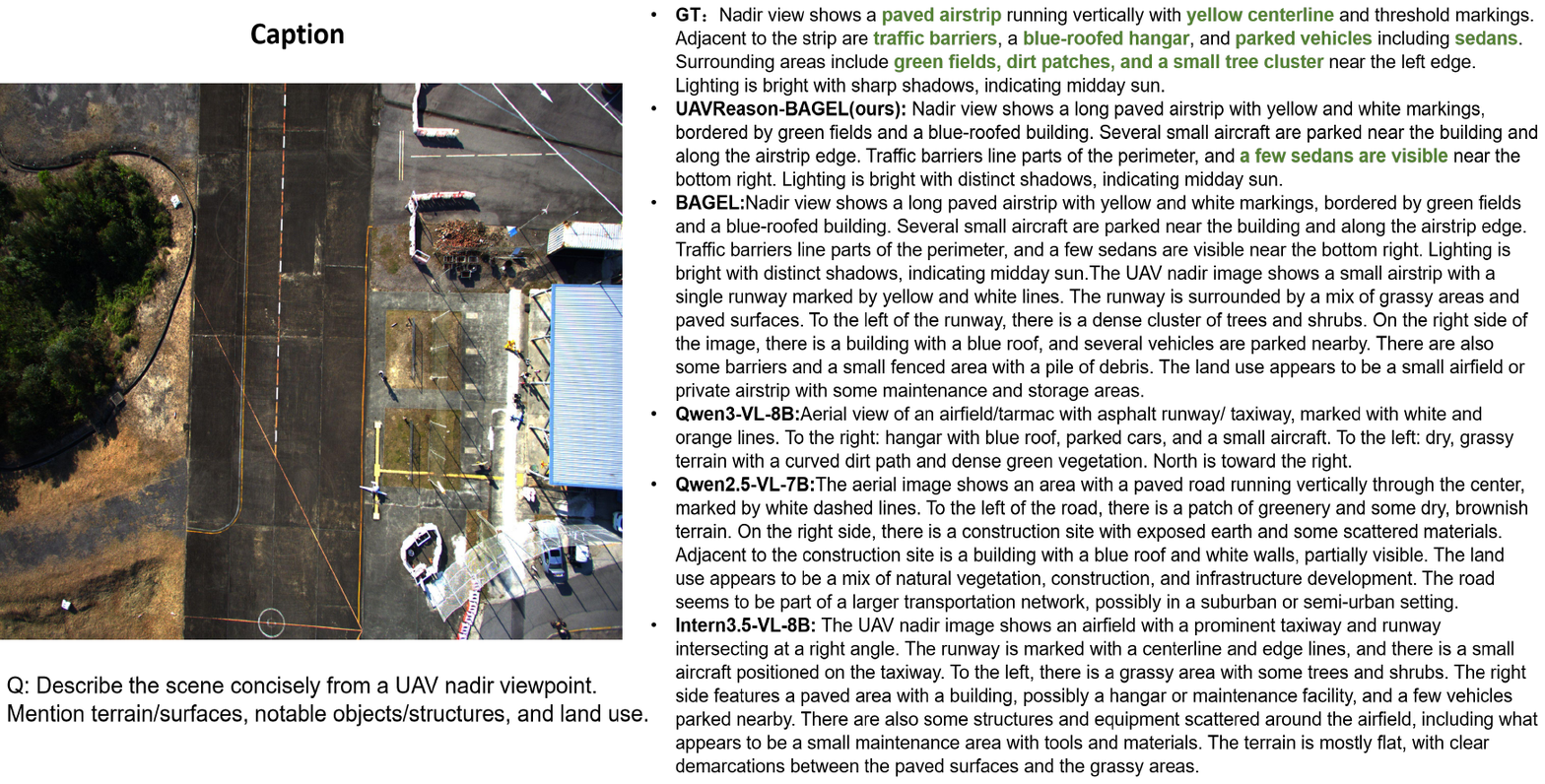}
  \vspace{-1.3in} 
\caption{\textbf{Qualitative comparison on fine-grained scene description.} 
  On a nadir-view airstrip, \textbf{UAVReason-Bagel (Ours)} achieves superior \emph{semantic granularity}, correctly identifying the context (``airstrip'') and resolving specific instances (\eg, ``sedans''). 
  In contrast, general-domain VLMs exhibit severe semantic drift, often hallucinating non-existent scenes (\eg, ``construction site'') or resorting to coarse abstractions (generic ``vehicles'').}
  \label{fig:qualitative_captioning}
\end{figure*}

We further provide concrete examples in Figure~\ref{fig:qualitative_reasoning} and Figure~\ref{fig:qualitative_captioning} to visualize the limitations of general-domain VLMs in aerial scenarios.
Despite their strong general capabilities, models like Qwen2-VL~\cite{qwen2.5-vl} and InternVL~\cite{internvl-2.5} exhibit unreliable grounding when facing the unique challenges of UAV imagery.
First, in VQA tasks, they struggle with \emph{scale and density}, frequently hallucinating counts for tiny objects like traffic barriers (Fig.~\ref{fig:qualitative_reasoning}, Left).
Second, they lack \emph{viewpoint adaptability}, failing to map explicit orientation prompts (\eg, rotated compasses) to the nadir coordinate system, leading to incorrect navigation inferences (Fig.~\ref{fig:qualitative_reasoning}, Middle).
Third, regarding \emph{temporal dynamics}, baselines often misinterpret fine-grained geometric changes, such as hallucinating an ``approaching'' motion when objects are clearly separating (Fig.~\ref{fig:qualitative_reasoning}, Right).
Finally, this grounding gap extends to \emph{scene description} (Fig.~\ref{fig:qualitative_captioning}). While baselines often resort to coarse abstractions (\eg, labeling specific sedans merely as ``vehicles'') or suffer from semantic drift (\eg, Qwen2.5-VL misidentifying an airstrip as a ``construction site''), UAVReason-Bagel preserves \emph{instance-level granularity}. It correctly resolves specific categories like ``sedans'' and captures the precise land use, confirming that our unified training strategy effectively mitigates domain-specific hallucinations across both reasoning and generation tasks.


\begin{figure*}[t!]
  \centering
  \includegraphics[width=\linewidth]{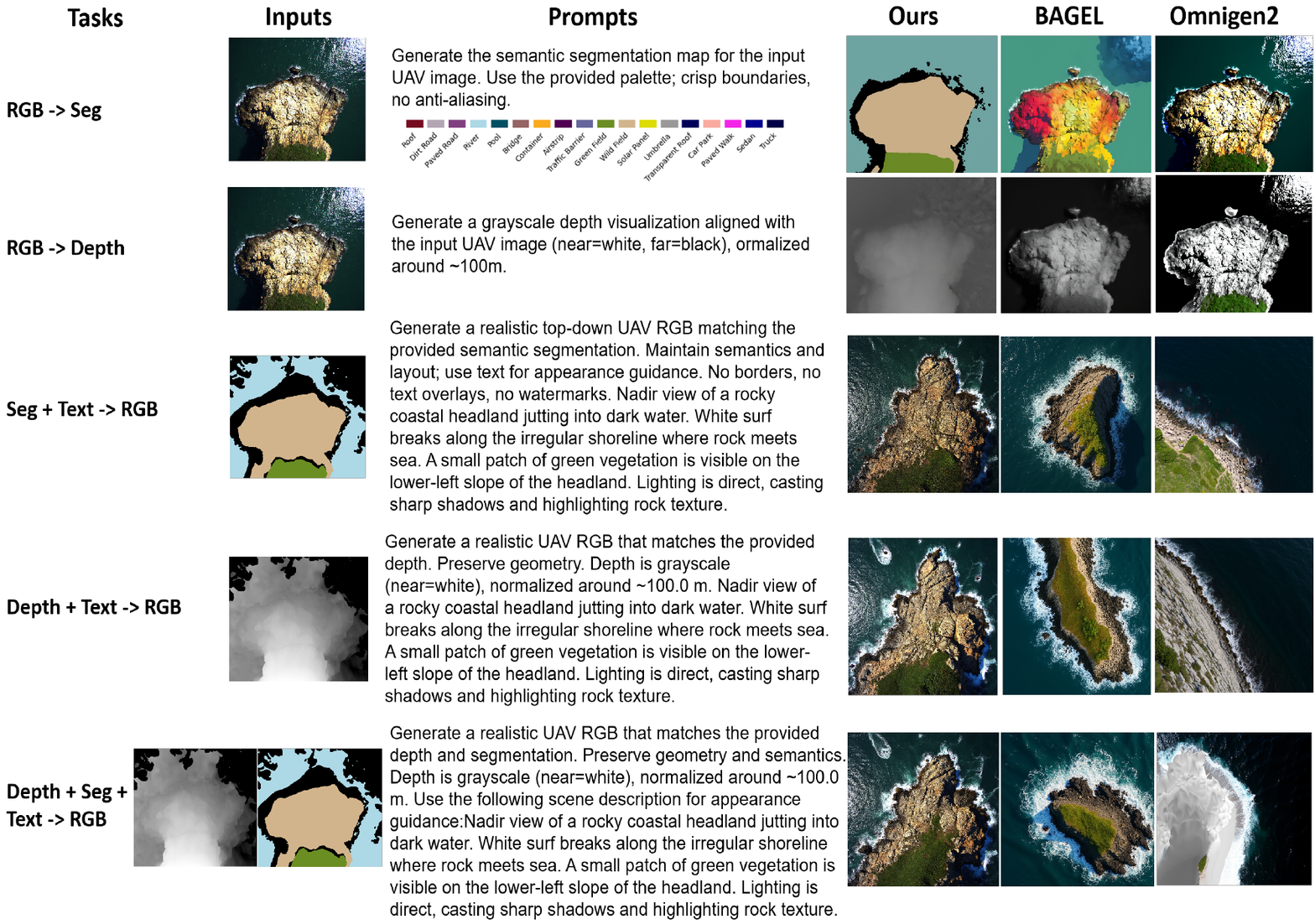}
  \vspace{-0.6cm} 
  \caption{\textbf{Qualitative comparison of dense perception and conditional synthesis.} 
  We visualize predictions from \textbf{UAVReason-Bagel} versus general-domain baselines (BAGEL, OmniGen2).
  In \textbf{perception tasks} (Top rows), our model recovers fine-grained object boundaries and smooth depth gradients, whereas baselines exhibit significant noise and semantic errors.
  In \textbf{conditional synthesis} (Bottom rows), our model generates photorealistic textures that strictly adhere to the input geometric layouts (depth/segmentation). In contrast, baselines suffer from severe domain shift, resulting in structural hallucinations and failure to follow the nadir-view control signals.}
  \label{fig:vis_gen1}
\end{figure*}

\subsection{Dense Perception and Generation Qualitative Analysis}
\label{sec:qual:gen}
\begin{figure*}[t!]
  \centering
  \includegraphics[width=\linewidth]{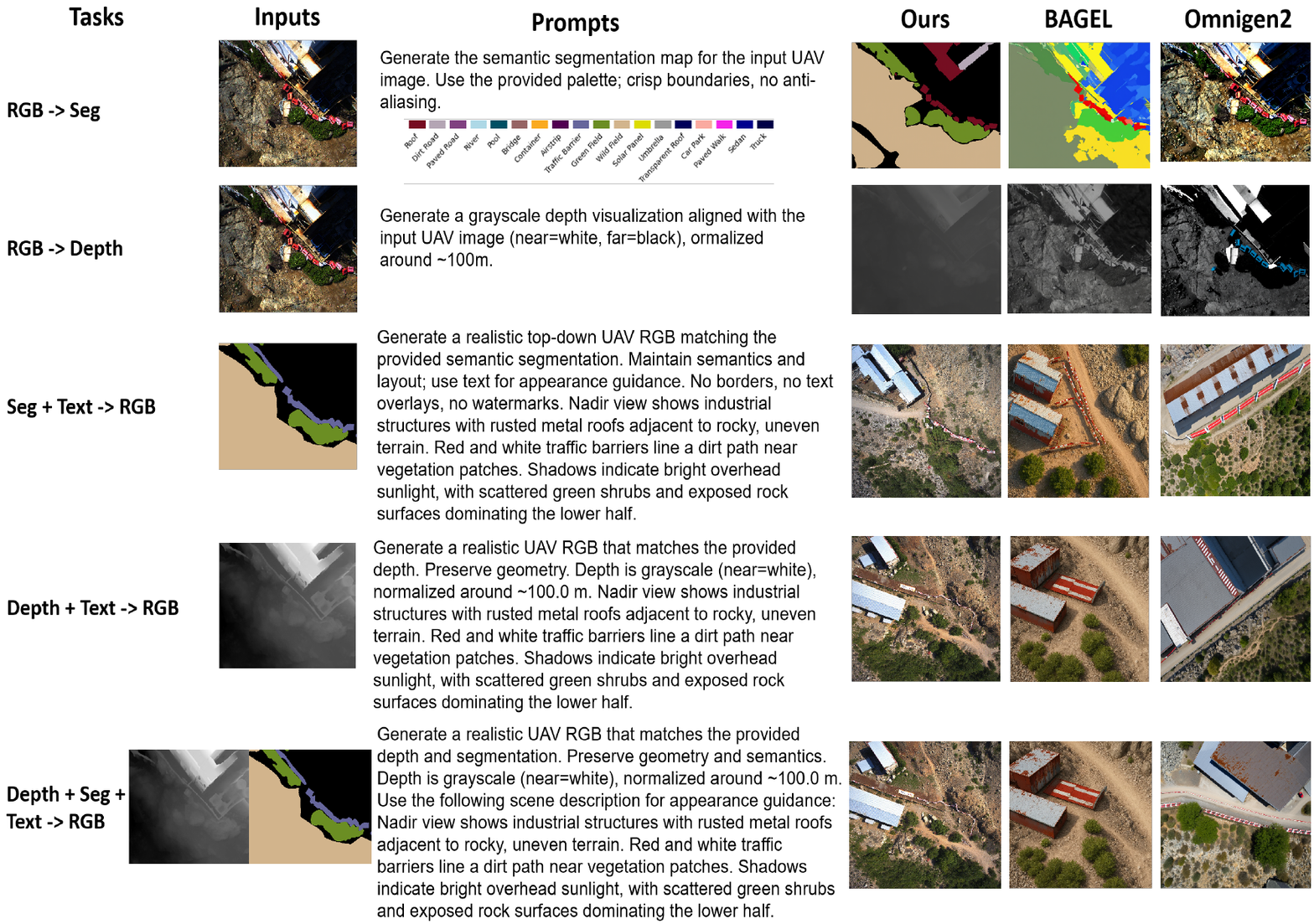}
  \vspace{-0.6cm} 
  \caption{\textbf{Qualitative comparison of dense perception and conditional synthesis.} 
  We visualize predictions from \textbf{UAVReason-Bagel} versus general-domain baselines (BAGEL, OmniGen2).
  In \textbf{perception tasks} (Top rows), our model recovers fine-grained object boundaries and smooth depth gradients, whereas baselines exhibit significant noise and semantic errors.
  In \textbf{conditional synthesis} (Bottom rows), our model generates photorealistic textures that strictly adhere to the input geometric layouts (depth/segmentation). In contrast, baselines suffer from severe domain shift, resulting in structural hallucinations and failure to follow the nadir-view control signals.}
  \label{fig:vis_gen2}
\end{figure*}

We provide a visual comparison in Figure~\ref{fig:vis_gen1} and Figure~\ref{fig:vis_gen2} to intuitively demonstrate the superiority of our unified approach.
\textbf{(1) Dense Perception:} As shown in the top rows, general-domain models (OmniGen2, BAGEL) struggle with the nadir viewpoint, often producing fragmented segmentation masks and noisy depth estimations that fail to capture the flat, layout-centric nature of UAV scenes. In contrast, UAVReason-Bagel makes clear, semantically coherent predictions with high structural fidelity.
\textbf{(2) Conditional Synthesis:} The bottom rows highlight a critical advantage in controllability. When generating RGB images from geometric conditions (Depth/Seg), baselines frequently hallucinate non-existent objects or distort the landscape, leading to weak alignment between the control signal and the generated output. Our model, however, maintains strict \emph{geometric adherence}, synthesizing realistic textures that perfectly align with the provided layout. This qualitative evidence validates our hypothesis that unified training on UAVReason effectively bridges the domain gap, mitigating the ``semantic floating'' issue prevalent in existing VLMs.

\section{Prompt Engineering Details}
\label{sec:appendix_prompts}
To ensure the generated Question-Answer (QA) pairs are strictly grounded in pixel-space and maintain semantic consistency, we employ a coarse-to-fine prompting pipeline. Furthermore, to guarantee the accuracy of answers, we utilize a Program-Aided approach where answers are derived from computational facts rather than direct LLM inference. The pipeline consists of three phases.
\paragraph{Hierarchical Image Enrichment}
\label{sec:enrichment}
As shown in Figure~\ref{fig:prompt_enrichment}, we first transform the raw visual data into a structured textual representation in two steps: Global Scene Parsing and Fine-grained Instance Reasoning.

\paragraph{Constraint-based QA Generation}
We prompt the LLM to generate diverse QA pairs based on the structured scene graph, balancing between single-frame cognition and multi-frame temporal reasoning, shown in Figure~\ref{fig:prompt_qa_generation}.

\begin{figure*}[t]
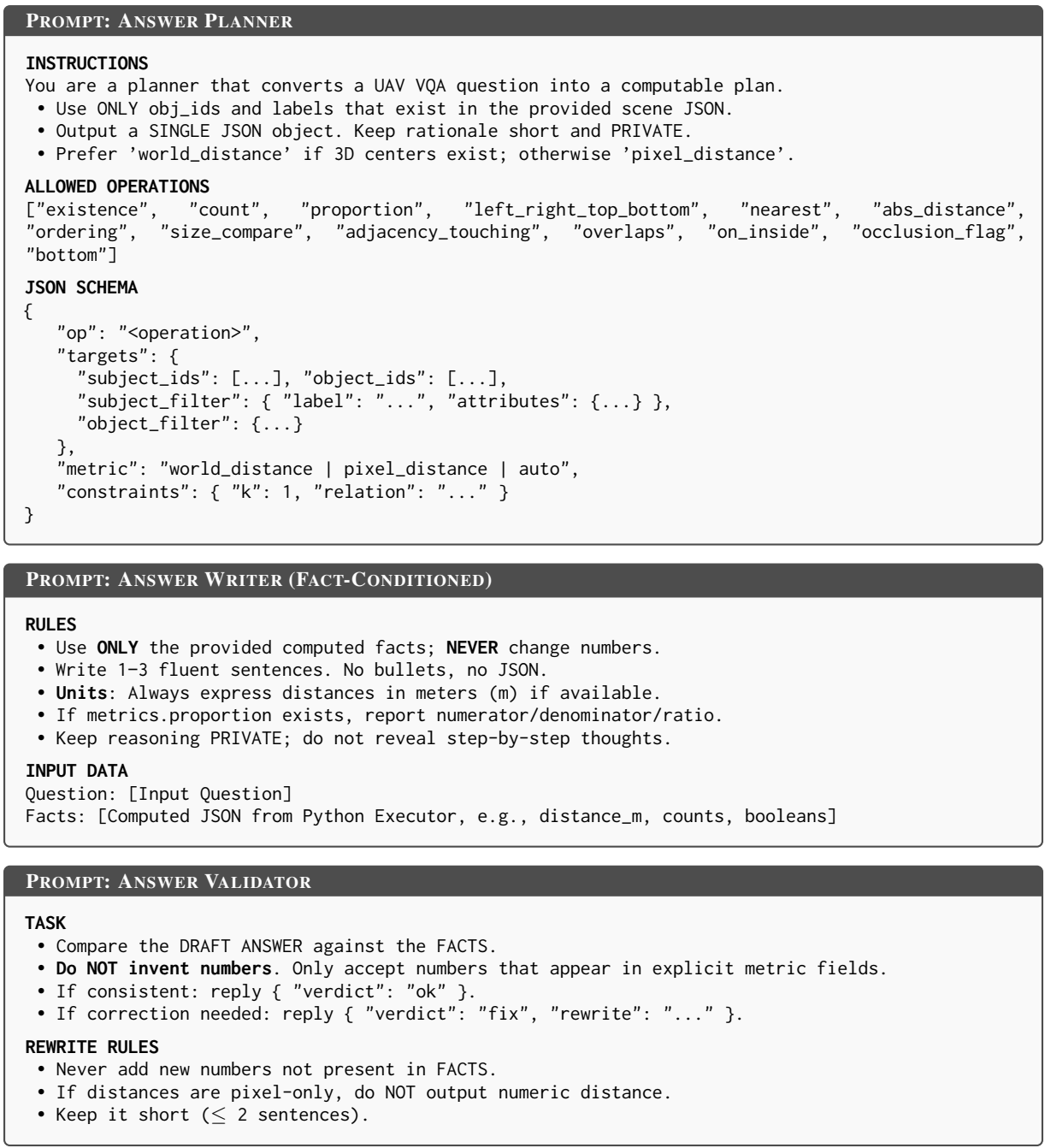

    \centering
    \begin{promptbox}{Prompt: Answer Planner}
\textbf{INSTRUCTIONS}\\
You are a planner that converts a UAV VQA question into a computable plan.
\begin{itemize}[leftmargin=1.5em, nosep]
    \item Use ONLY \texttt{obj\_ids} and labels that exist in the provided scene JSON.
    \item Output a SINGLE JSON object. Keep rationale short and PRIVATE.
    \item Prefer \texttt{'world\_distance'} if 3D centers exist; otherwise \texttt{'pixel\_distance'}.
\end{itemize}

\vspace{0.5em}
\textbf{ALLOWED OPERATIONS}\\
{[}"existence", "count", "proportion", "left\_right\_top\_bottom", "nearest", "abs\_distance", "ordering", "size\_compare", "adjacency\_touching", "overlaps", "on\_inside", "occlusion\_flag", "bottom"{]}

\vspace{0.5em}
\textbf{JSON SCHEMA}\\
\{ \\
\hspace*{1em} "op": "<operation>", \\
\hspace*{1em} "targets": \{ \\
\hspace*{2em} "subject\_ids": [...], "object\_ids": [...], \\
\hspace*{2em} "subject\_filter": \{ "label": "...", "attributes": \{...\} \}, \\
\hspace*{2em} "object\_filter": \{...\} \\
\hspace*{1em} \}, \\
\hspace*{1em} "metric": "world\_distance | pixel\_distance | auto", \\
\hspace*{1em} "constraints": \{ "k": 1, "relation": "..." \} \\
\}
    \end{promptbox}
    \vspace{-0.2cm}

    \begin{promptbox}{Prompt: Answer Writer (Fact-Conditioned)}
\textbf{RULES}
\begin{itemize}[leftmargin=1.5em, nosep]
    \item Use \textbf{ONLY} the provided computed facts; \textbf{NEVER} change numbers.
    \item Write 1--3 fluent sentences. No bullets, no JSON.
    \item \textbf{Units}: Always express distances in meters (m) if available.
    \item If \texttt{metrics.proportion} exists, report numerator/denominator/ratio.
    \item Keep reasoning PRIVATE; do not reveal step-by-step thoughts.
\end{itemize}

\vspace{0.5em}
\textbf{INPUT DATA}\\
Question: [Input Question] \\
Facts: [Computed JSON from Python Executor, e.g., distance\_m, counts, booleans]
    \end{promptbox}
    \vspace{-0.2cm}
    
    \begin{promptbox}{Prompt: Answer Validator}
\textbf{TASK}
\begin{itemize}[leftmargin=1.5em, nosep]
    \item Compare the DRAFT ANSWER against the FACTS.
    \item \textbf{Do NOT invent numbers}. Only accept numbers that appear in explicit metric fields.
    \item If consistent: reply \{ "verdict": "ok" \}.
    \item If correction needed: reply \{ "verdict": "fix", "rewrite": "..." \}.
\end{itemize}

\vspace{0.5em}
\textbf{REWRITE RULES}
\begin{itemize}[leftmargin=1.5em, nosep]
    \item Never add new numbers not present in FACTS.
    \item If distances are pixel-only, do NOT output numeric distance.
    \item Keep it short ($\le$ 2 sentences).
\end{itemize}
    \end{promptbox}

    \vspace{-0.2cm}
    \caption{\textbf{Prompts for the Program-Aided Answer Generation.} The pipeline consists of three stages: the \textbf{Planner} parses the natural language question into a structured JSON query; the \textbf{Writer} synthesizes a fluent answer based on ground-truth facts computed by the deterministic executor; and the \textbf{Validator} performs a final sanity check to ensure no numerical hallucinations are included.}
    \label{fig:prompt_answer}
\end{figure*}

\paragraph{Program-Aided Answer Synthesis}
\noindent To mitigate hallucination in language-based reasoning, we adopt a \textit{Plan-Execute-Verify} paradigm (see Figure~\ref{fig:prompt_answer}). The LLM first acts as a Planner to generate a symbolic program. A Python executor then computes precise metrics (e.g., Euclidean distance in meters) from Ground Truth data. Finally, a Writer synthesizes the natural language answer, checked by a Validator.
\clearpage

\begin{figure*}[t]
    \centering
    \vspace{-1.5em}
    \begin{promptbox}{Prompt: Step-1 Global Scene Parsing}
\textbf{INSTRUCTIONS (STEP-1 ONLY)}
You are a professional UAV nadir/near-nadir image-to-text annotator.
\begin{itemize}[leftmargin=1.5em, nosep]
    \item Based on the aerial image and the CANDIDATE instances list, output STRICT JSON with ONLY: "version", "image\_size", "scene".
    \item DO NOT output "instances" or "relations" in this step.
\end{itemize}

\vspace{0.5em}
\textbf{HARD RULES}
\begin{itemize}[leftmargin=1.5em, nosep]
    \item Image coords: origin top-left; $x \to \text{right}$, $y \to \text{down}$.
    \item \textbf{Ontology}: Do NOT invent new labels; ontology is fixed.
    \item \textbf{Language}: Concise, factual, pixel-grounded (nadir cues). Avoid hedging.
    \item \textbf{Directions}: Only left/right/top/bottom, near/far; never N/E/S/W.
\end{itemize}

\vspace{0.5em}
\textbf{OUTPUT JSON SCHEMA}\\
\{ \\
\hspace*{1em} "version": "<model\_version>", \\
\hspace*{1em} "image\_size": \{ "width": [W], "height": [H] \}, \\
\hspace*{1em} "scene": \{ \\
\hspace*{2em} "scene\_type": "residential | industrial | ...", \\
\hspace*{2em} "global\_caption": "3-4 sentences with nadir cues...", \\
\hspace*{2em} "main\_surfaces": ["paved\_motor\_road", ...], \\
\hspace*{2em} "time\_of\_day": "day | dusk | night | unknown" \\
\hspace*{1em} \} \\
\}
    \end{promptbox}
    \vspace{-0.2cm}

    \begin{promptbox}{Prompt: Step-2 Instance Attribute \& Relation Reasoning}
\textbf{INSTRUCTIONS (STEP-2 ONLY)}
You are a professional UAV nadir/near-nadir image-to-text annotator.
\begin{itemize}[leftmargin=1.5em, nosep]
    \item Using ONLY the provided STEP-1 scene JSON and ACCEPTED\_PROPOSALS, return STRICT JSON with: "by\_obj" and "relations".
    \item DO NOT return bbox/label/src\_id in "by\_obj".
\end{itemize}

\vspace{0.5em}
\textbf{by\_obj VALUE SCHEMA}\\
\{ \\
\hspace*{1em} "attributes": \{ \\
\hspace*{2em} "color": "...", "condition": "intact | damaged", \\
\hspace*{2em} "state": "stationary | moving", "occluded": true | false \\
\hspace*{1em} \}, \\
\hspace*{1em} "attribute\_sentence": "1--2 factual sentences...", \\
\hspace*{1em} "spatial\_context\_sentence": "1--2 sentences using relative directions...", \\
\hspace*{1em} "evidence": "$\le$ 10 words" \\
\}

\vspace{0.5em}
\textbf{HARD RULES}
\begin{itemize}[leftmargin=1.5em, nosep]
    \item Produce an entry in "by\_obj" for \textbf{EVERY} obj\_id given.
    \item Use only relative directions (left/right/top/bottom).
    \item Relations must reference existing obj\_ids and be pixel-grounded.
\end{itemize}
    \end{promptbox}

    \vspace{-0.2cm}
    \caption{\textbf{Prompts for the Image Enrichment phase.} Step-1 verifies global layout, while Step-2 fills fine-grained attributes and spatial relations.}
    \label{fig:prompt_enrichment}
\end{figure*}
\begin{figure*}[t]
    \centering
    \small 
    
    \begin{promptbox}{Prompt: Single-Frame VQA Generation}
\textbf{ROLE}: You are a professional UAV nadir/near-nadir VQA author.

\textbf{GOAL}
\begin{itemize}[leftmargin=1.2em, nosep]
    \item Design \textbf{EXACTLY 9 verifiable questions} using ONLY the provided scene JSON.
    \item Output JSON ONLY with: \texttt{\{"generated\_questions":[ ... ]\}}.
\end{itemize}

\vspace{0.3em}
\textbf{CATEGORY \& COUNT (STRICT)}
\begin{itemize}[leftmargin=1.2em, nosep]
    \item \texttt{Common Scenes}: "Existence", "Count", "Proportion", "Attribute"
    \item \texttt{Spatial Reasoning}: "Adjacency", "Distance", "Direction", "Occlusion"
    \item \texttt{CoT Multi-Hop}: "Composition", "Constraint", "Filter-Then-Measure"
\end{itemize}

\vspace{0.3em}
\textbf{STRONG CONSTRAINTS}
\begin{itemize}[leftmargin=1.2em, nosep]
    \item Each question MUST reference an existing \texttt{obj\_id} or \texttt{label}.
    \item Keep each question $\le$ 22 words; concise and unambiguous.
    \item Favor \textbf{pixel-space geometry}. Do not use meters unless explicitly provided.
\end{itemize}
    \end{promptbox}

    \vspace{0.15cm} 

    \begin{promptbox}{Prompt: Temporal (Multi-Frame) VQA Generation}
\textbf{ROLE}: You are a professional UAV nadir/near-nadir VQA author.

\textbf{GOAL}
\begin{itemize}[leftmargin=1.2em, nosep]
    \item Design \textbf{EXACTLY 3 verifiable questions} using ONLY the provided TWO-FRAME JSON.
    \item Categories: \texttt{"Identity"}, \texttt{"DistanceChange"}, \texttt{"Speed"}, \texttt{"RelationChange"}, \texttt{"Approach/Separate"}.
\end{itemize}

\vspace{0.3em}
\textbf{TEMPORAL RULES}
\begin{itemize}[leftmargin=1.2em, nosep]
    \item EVERY question MUST cite \textbf{BOTH frames} (@t0 and @t1).
    \item Prefer using "tracks" (\texttt{trk\#k}) to refer to the same instance across frames.
    \item \textit{*Note: Temporal reasoning requires strict object grounding across timestamps.}
\end{itemize}
    \end{promptbox}

    \vspace{-0.2cm}
    \caption{\textbf{Prompts for Question Generation.} The top block instructs the model to cover diverse cognitive tasks (e.g., counting, spatial reasoning) on single frames. The bottom block enforces cross-frame tracking and dynamic state analysis for temporal understanding.}
    \label{fig:prompt_qa_generation}
\end{figure*}

\end{document}